\documentclass[10pt,twocolumn,letterpaper]{article}

\usepackage{cvpr}              % To produce the CAMERA-READY version
\usepackage{epsfig}
\usepackage{graphicx}
\usepackage{amsmath}
\usepackage{amsthm}
\usepackage{color}
\usepackage{booktabs}
\usepackage{multirow}
\usepackage{adjustbox}

\usepackage{algorithmic}
\usepackage{algorithm}
\usepackage{marvosym}
\usepackage{float}
\definecolor{cvprblue}{rgb}{0.21,0.49,0.74}
\usepackage[pagebackref,breaklinks,colorlinks,allcolors=cvprblue, urlcolor=red]{hyperref}
\def\paperID{***} % *** Enter the Paper ID here
\def\confName{CVPR}
\def\confYear{2025}

\newcommand\blfootnote[1]{%
	\begingroup
	\renewcommand\thefootnote{}\footnote{#1}%
	\addtocounter{footnote}{-1}%
	\endgroup
}

\title{Learning Affine Correspondences by Integrating Geometric Constraints}

\author{
Pengju Sun\textsuperscript{1, 2}\hspace{0.2em}  Banglei Guan\textsuperscript{1, 2(\Letter)}\hspace{0.2em}  Zhenbao Yu\textsuperscript{1, 2} \hspace{0.2em} Yang Shang\textsuperscript{1, 2}\hspace{0.2em}  Qifeng Yu\textsuperscript{1, 2}\hspace{0.2em}  Daniel Barath\textsuperscript{3, 4}\\[2mm]
	{$^{1}$College of Aerospace Science and Engineering,}  { National University of Defense Technology, China.} \\[0.5mm]
    {$^{2}$   Hunan Provincial Key Laboratory of Image Measurement and Vision Navigation, China.} \\[0.5mm]
	{$^{3}$ETH Zurich, Switzerland.}
    {$^{4}$ HUN-REN SZTAKI, Hungary.}
}

\begin{document}
\maketitle

\begin{abstract}
\indent Affine correspondences have received significant attention due to their benefits in tasks like image matching and pose estimation. Existing methods for extracting affine correspondences still have many limitations in terms of performance; thus, exploring a new paradigm is crucial. In this paper, we present a new pipeline designed for extracting accurate affine correspondences by integrating dense matching and geometric constraints. Specifically, a novel extraction framework is introduced, with the aid of dense matching and a novel keypoint scale and orientation estimator. For this purpose, we propose loss functions based on geometric constraints, which can effectively improve accuracy by supervising neural networks to learn feature geometry. The experimental show that the accuracy and robustness of our method outperform the existing ones in image matching tasks. To further demonstrate the effectiveness of the proposed method, we applied it to relative pose estimation. Affine correspondences extracted by our method lead to more accurate poses than the baselines on a range of real-world datasets. The code is available at \href{https://github.com/stilcrad/DenseAffine}{https://github.com/stilcrad/DenseAffine}.

\end{abstract}

\blfootnote{\Letter: Corresponding author. \\ Email:\texttt{ \{sunpengju23, guanbanglei12, \\ shangyang1977, yuqifeng\}@nudt.edu.cn}\\ \texttt{zhenbaoyu@whu.edu.cn}  \texttt{dbarath@ethz.ch} }
\section{Introduction}
\label{Introduction}

\noindent\hspace{1em} In computer vision, image matching and geometric estimation stand as fundamental problems, playing crucial roles in domains ranging from autonomous driving to robotics~\cite{barath_polic_förstner_sattler_pajdla_kukelova_2020,guan_iccv_2021,GeoDesc}. Affine correspondences (ACs) have attracted significant attention in recent years, due to their ability to provide valuable insights into the underlying 3D geometry of the surrounding environment~\cite{2ac2018barath,guan_eccv_2022,Barth2023OnMS}.
Notably, previous research has demonstrated the efficacy of affine correspondences in tasks such as homography, epipolar geometry, and focal length estimation~\cite{barath_toth_hajder_2017,raposo_barreto_2016,Hruby2024, yuzhenbao24}. 

\begin{figure}[tpb]
    \centering
    \begin{minipage}[b]{\linewidth}
        \subfloat[LoFTR]{
            \includegraphics[width=0.3\linewidth]{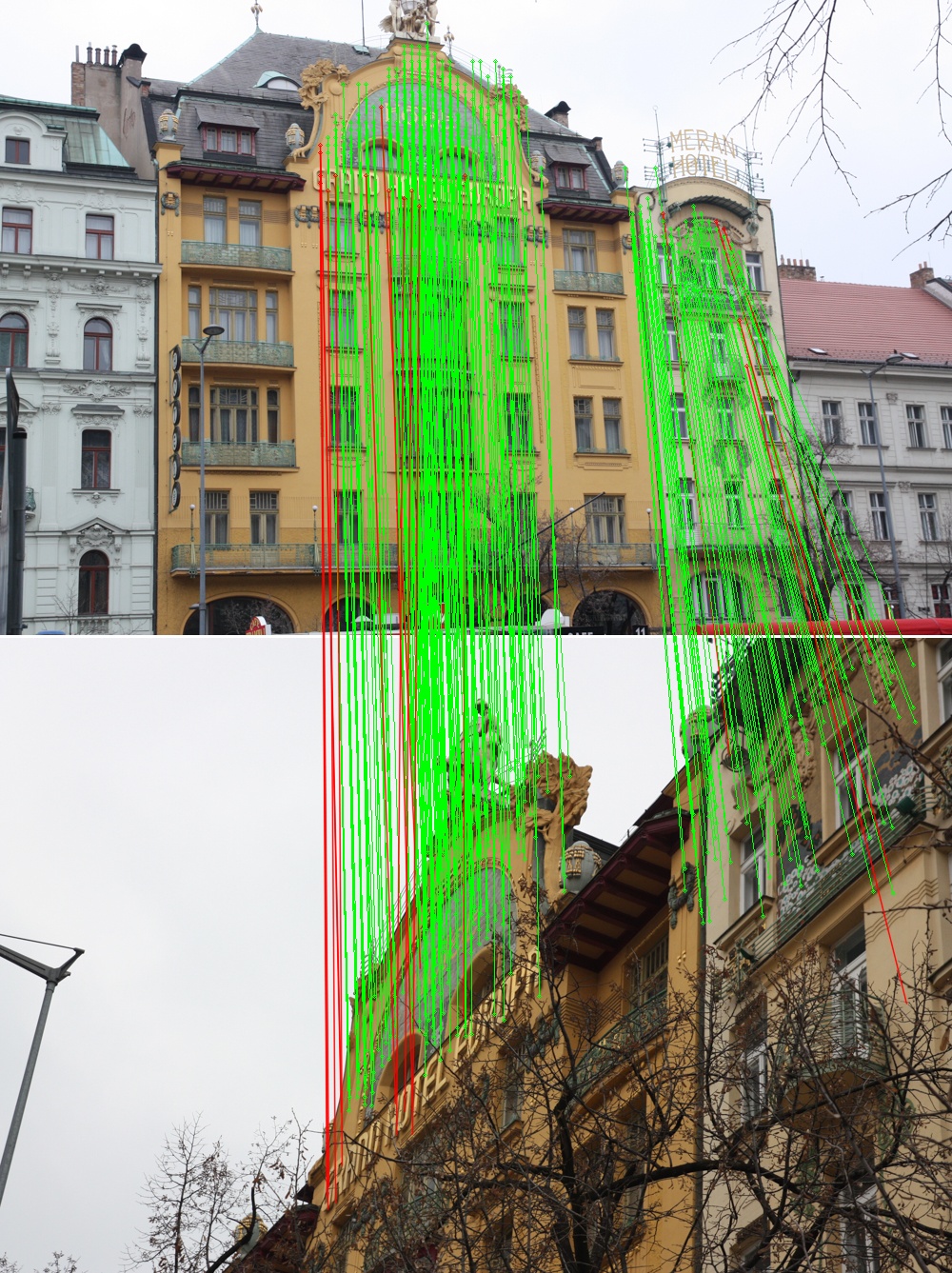}
        }\hfill
        \subfloat[DKM]{
            \includegraphics[width=0.3\linewidth]{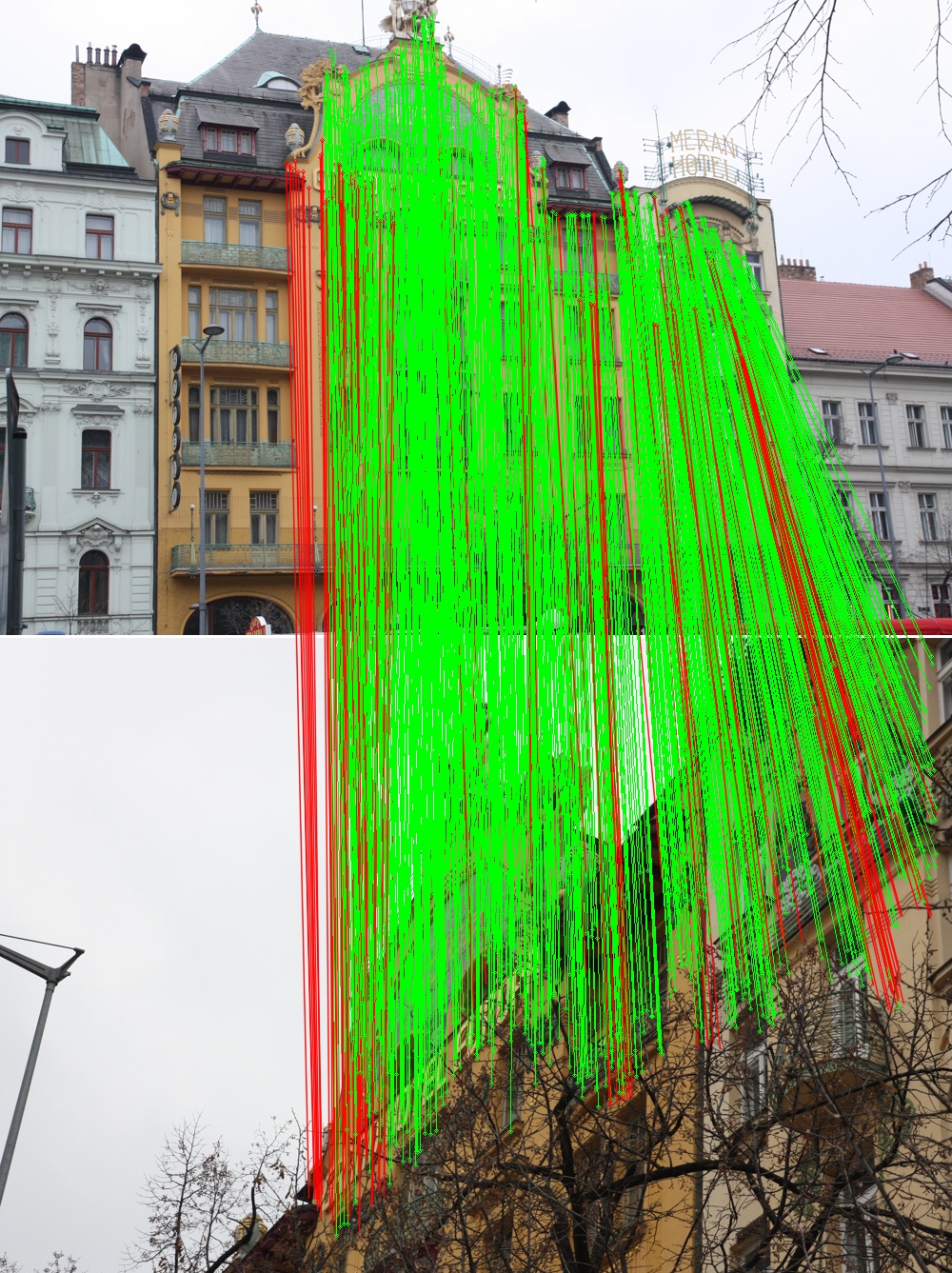}
        }\hfill
        \subfloat[Ours]{
        \includegraphics[width=0.3\linewidth]{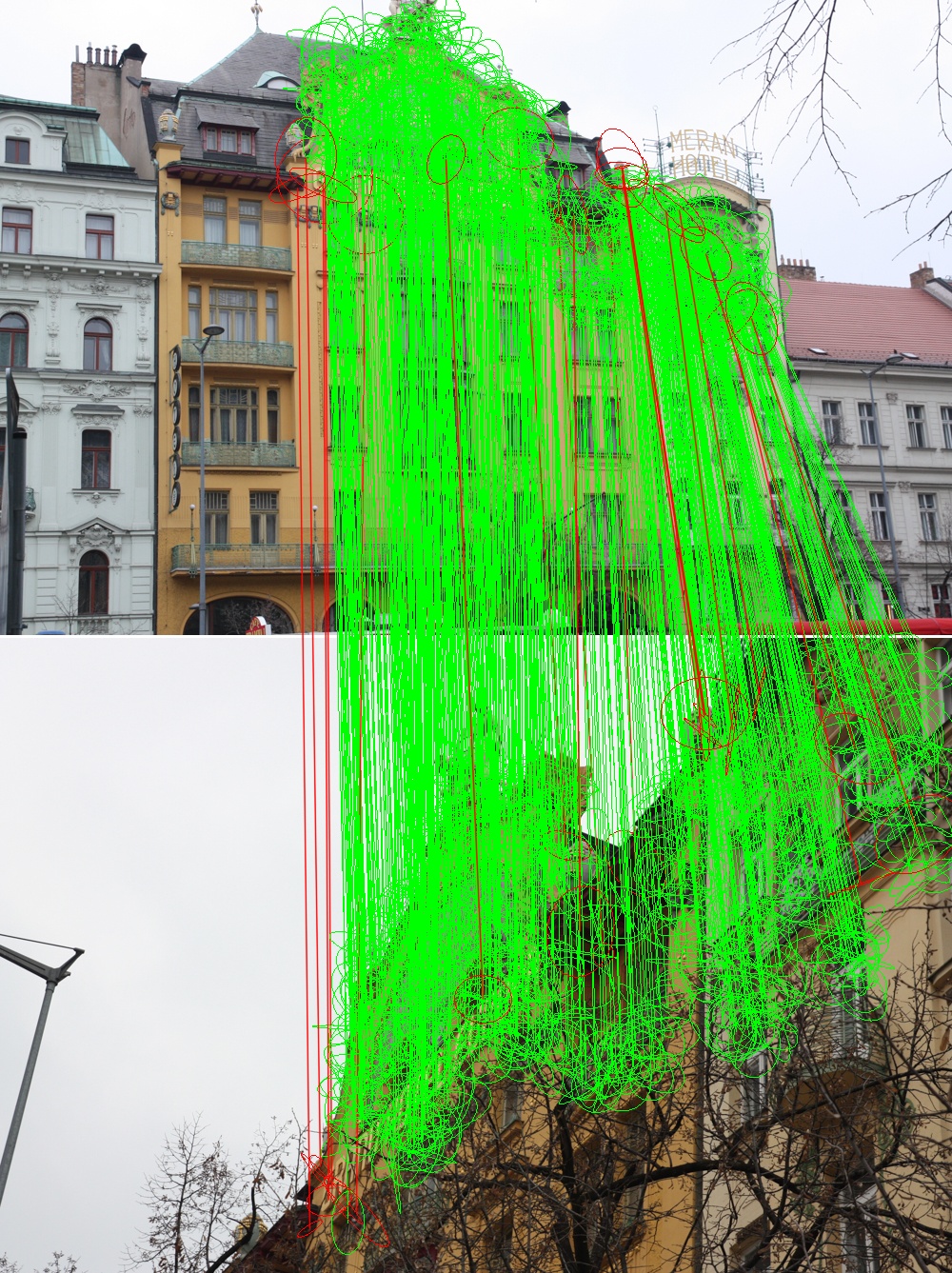}
        }\hfill
    \end{minipage}
    \caption{Image matching with large viewpoint change. Correct matches are green, and incorrect ones red.
        Our method leads to more correct matches than the LoFTR~\cite{sun2021loftr} and DKM~\cite{DKM}. }
    \label{fig:1}
    \vspace{-0.4cm}
\end{figure}

Affine correspondences offer distinct advantages in addressing fundamental challenges in visual perception, particularly in image-matching and relative pose estimation tasks, owing to its capacity to encode higher-order information about the scene geometry ~\cite{barath_polic_förstner_sattler_pajdla_kukelova_2020,guan_iccv_2021,eichhardt_barath_2019}.
By locally approximating the image deformation caused by changes in camera pose through affine mapping, geometric information about corresponding local regions are obtained~\cite{MISHKIN201581,Guan_1ac_relative}.
These approximations enhance the robustness of affine covariants for matching and recognition tasks~\cite{mikolajczyk_mikolajczyk_2004}.
It is a valuable property for matching planar surfaces in the presence of extreme viewpoint changes, improving the reliability of wide baseline image matching~\cite{affnet}.
Moreover, affine correspondences are used to estimate complex geometric relationships between images, such as essential matrices, outperforming the results when relying solely on point correspondences~\cite{2ac2018barath,barath_polic_förstner_sattler_pajdla_kukelova_2020}. Benefiting from the smaller sample sizes
required, robust homography and relative pose estimation is significantly faster than when using the point-based solvers while leading to more accurate results~\cite{2ac2018barath,barath_polic_förstner_sattler_pajdla_kukelova_2020}. 
Through exploiting these informative affine correspondences, such algorithms can attain enhanced precision and reduced runtime ~\cite{barath_polic_förstner_sattler_pajdla_kukelova_2020}. 
The reduction in the number of matches needed to estimate a model, \eg, homography, leads to advantages, like reduced computational complexity and better efficiency of the outlier removal process~\cite{Guan_1ac_relative,guan2021minimal,guan2023minimal}.

Obtaining high-quality affine correspondences in real-world scenarios remains an open problem~\cite{Rodriguez_Facciolo_von,Barth2023OnMS}. There are many limitations to existing methods, such as the limited quantity and accuracy. 
These drawbacks arise from many of these methods that use detector-based techniques and do not fully exploit geometric constraints. 
Existing extractors that use sparse detectors perform poorly with repetitive and weak textures~\cite{Vedaldi2010VlfeatAO,affnet}. %, which rely on local repetitive features. 
View-synthesis-based methods, such as ASIFT~\cite{ASIFT}, rely not only on the detector, but also involve computationally expensive image transformations.
These limitations severely affect the use of affine correspondences in tasks related to geometric estimation~\cite{dai_jin_zhang_nguyen_2020}.

In recent years, dense matching with neural networks has been shown to effectively overcome the limitations imposed by traditional detection methods~\cite{sun2021loftr,DKM}, which rely on detected sparse keypoints~\cite{ASIFT,Vedaldi2010VlfeatAO,affnet}. 
Having a dense warp between the two images allows one to extract Aundant  accurate keypoints. Due to the use of global context, such methods also excel in weakly textured regions and repetitive structures. 
At the same time, geometric constraints are also proving their effectiveness in matching tasks~\cite{caps_2020}. 

This paper presents a novel pipeline  for robust affine correspondence estimation through synergistic integration of dense feature matching and geometric constraint optimization.
This approach allows us to extract a large number of accurate corespondences even between images with large viewpoint changes, as shown in Fig.\ref{fig:1}.
In summary, the main contributions are as follows.
\begin{itemize}
    \item We propose a novel framework for estimating affine correspondences. By combining a dense matcher, geometric constraints, and a local affine transformation extractor using a soft scale and orientation estimator, The framework surpasses state-of-the-art in match number and accuracy.
    \item A novel affine transformation loss, represented by the Affine Sampson Distance, is introduced to further enhance the conformity of affine correspondences with the scene geometry. This approach enhances training supervision, improving affine correspondence quality.
    \item  We show that the proposed model is applicable to a range of matching tasks, producing high-quality affine correspondences and achieving state-of-the-art results on both image-matching and relative pose estimation benchmarks, indicating the robustness, performance, and applicability of the method in practical scenarios.
\end{itemize}

\section{Related Work}
\label{Section:Related Work}
\subsection{Image Matching}
\noindent\hspace{1em} Classical image matching involves three key steps: keypoint detection, descriptor extraction, and correspondence estimation~\cite{ jin_mishkin_mishchuk_matas_fua_yi_trulls_2021}. Keypoints are traditionally detected using scale pyramids with handcrafted response functions~\cite{lowe_2004}. Feature matching relies on optimizing metrics like Sum of Squared Differences or correlation.~\cite{XU2024102344}. Examples of such functions include the Hessian~\cite{2001hessian_affine}, Harris~\cite{Harris1988ACC}, Difference of Gaussians (DoG)~\cite{lowe_2004}, as well as learned ones such as SOSNet~\cite{tian_yu_fan_wu_heijnen_balntas_2019}, Key.Net~\cite{laguna_riba_ponsa_mikolajczyk_2019}, SuperPoint~\cite{SuperPoint}, PoSFeat~\cite{posfeat}, and DKDNet~\cite{Gao2024DKDNet}. SuperPoint adopts a detector-based architecture similar to handcrafted methods and proposes a self-supervised training approach through homography adaptation, yielding improved performance. Subsequently, several algorithms have been devised based on these paradigms~\cite{r2d2,disk,edstedt2024dedode}. DKDNet integrates a dynamic keypoint feature learning module and a guided heatmap activator to enhance the performance of keypoint detection~\cite{Gao2024DKDNet}. These methods depend on the efficacy of the designed detectors, encountering limitations in scenarios with repetitive structures or weak textures.
	
Concurrent with detector-based matching, another line of works \cite{ SuperPoint, d2net,r2d2} focus on generating matches directly from raw images, where richer context can be leveraged and the keypoint detection step skipped. They execute global matching uniformly across the image grid at a coarse scale and extract matches via mutual-nearest neighbors or optimal transport~\cite{NCnet,Patch2Pix,sun2021loftr,EfficientLoFTR}. In contrast to detector-free methods, dense methods generate a dense warp. This warp is typically predicted by regression based on the global 4D-correlation volume~\cite{PDC-Net+}. DKM~\cite{DKM} introduces a dense kernelized matching approach that significantly improves two-view estimation. Building on this, RoMa~\cite{roma} represents a significant advancement in dense feature matching by applying a Markov chain framework to analyze and improve the matching process. GIM~\cite{Shen2024GIMLG} is a self-training method for matching using internet videos. Dense approaches have the capability to estimate matching pixel pairs, providing a foundation for obtaining precise affine correspondences.

\subsection{Affine Correspondence Estimation }
\noindent\hspace{1em} An affine correspondence consists of a pair of points along with the corresponding local affine transformation that maps the neighborhood of a point in the first image to its counterpart in the second image~\cite{raposo_barreto_2016,2ac2018barath}. Affine covariant detectors are commonly used to estimate affine transformations~\cite{2001hessian_affine}. These detectors typically fall into three categories. The first category includes methods like Maximally Stable Extremal Regions (MSER)~\cite{mser}, which directly estimate full local affine transformations from image regions. 
The second category consists of detectors such as Harris-Affine~\cite{2002An} and Hessian-Affine~\cite{mikolajczyk_mikolajczyk_2004}, which refine initial estimates using iterative methods such as Baumberg iteration~\cite{2001hessian_affine}, resulting in high-quality affinities. Some methods in the third category synthesize views related by affine transformations and then apply feature detectors to these synthetic images~\cite{ASIFT,mishkin2015mods}. Each point correspondence generates a local affinity, which is integrated with the synthetic view transformation to form the final affine feature.
	
In recent years, deep learning-based feature matching has shown significant advances in performance and robustness. AffNet~\cite{affnet} is proposed to demonstrate that repeatability is not enough. It learns local affine-covariant regions by optimizing a descriptor-based loss. 
AffNet outperforms prior methods in affine shape estimation and enhances the state-of-the-art in image retrieval. 
LOCATE~\cite{rodríguez_delon_morel_2018} incorporates local affine maps between corresponding keypoints, substantially improving the accuracy of local geometry estimation. AEU~\cite{dai_jin_zhang_nguyen_2020} is introduced to enhance feature matching accuracy by estimating relative affine transformations between features, making it more robust to disturbances.

Although these methods leverage neural networks to estimate local affine transformations and enhance image matching accuracy, their  still rely on sparse detectors. As a result, they inherit limitations such as sensitivity to low-textured regions and repetitive patterns. 
 
In contrast, our approach adopts a dense matching strategy, enabling accurate correspondences.

\begin{figure*}[htp]
    \centering
    \setlength{\abovedisplayskip}{3pt}
    \setlength{\belowdisplayskip}{3pt}
    \includegraphics[width=0.95\textwidth,trim={0.1cm 1.05cm 0.1cm 1.05cm},clip]
    {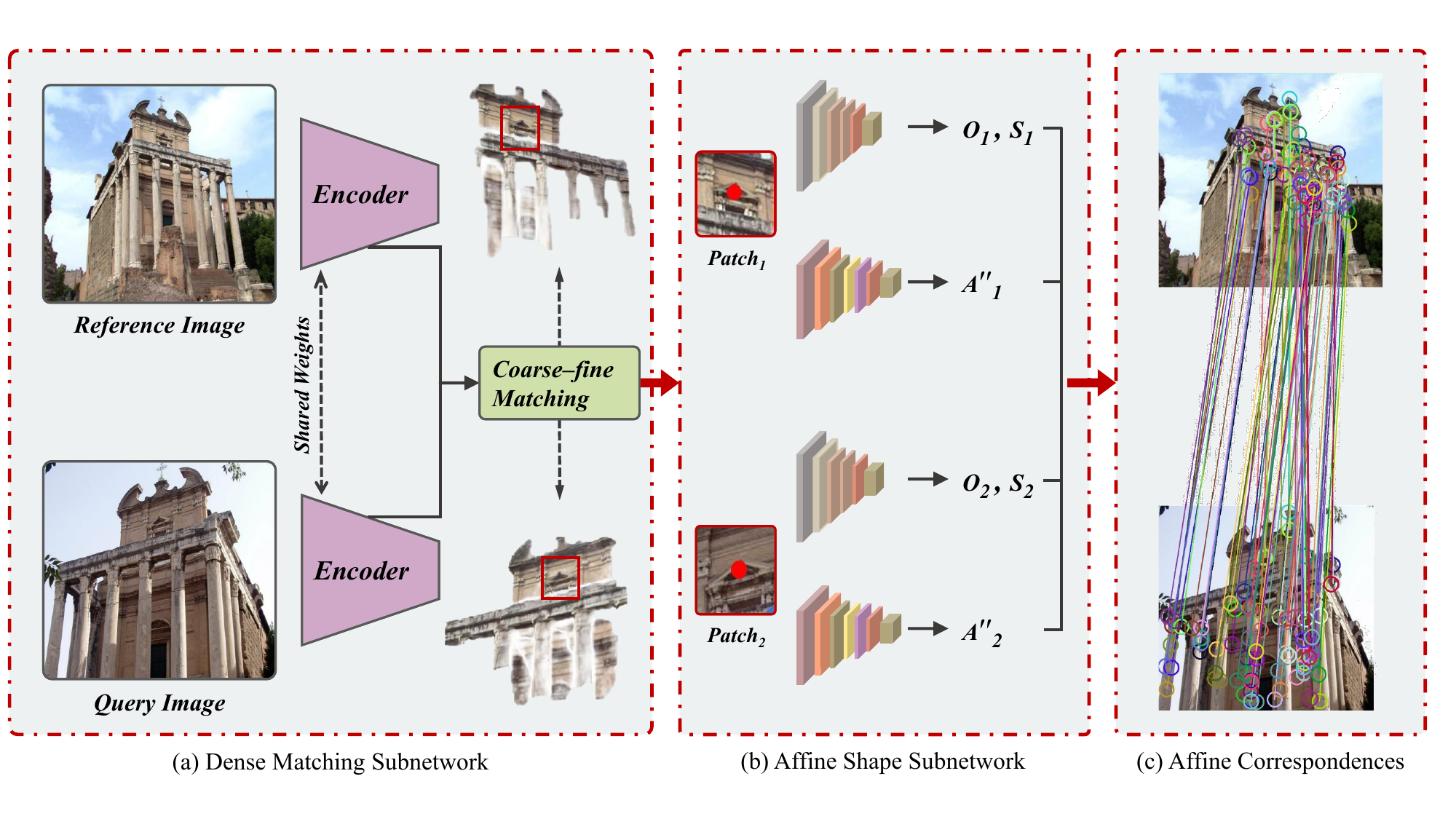} 
    \caption{The overview of our method. 
    (a) Abundant accurate point correspondences, encouraged to comply with epipolar constraints through the training loss, are obtained via a dense matching sub-network.
    (b) The second sub-network is used to estimate the orientation $\mathbf O_i$ and scale $\mathbf S_i$ of each patch and estimate the residual shape $\mathbf A_i''$. (c) Affine correspondences between the two images are estimated.  }
    \label{overview}
    \vspace{-1em}
\end{figure*}

\section{Proposed Method}
\label{Section: Method}
\noindent\hspace{1em} To obtain accurate affine correspondences, it is essential to ensure the precision of each component involved, including high-precision point correspondences and local affine transformations. In this section, a new framework is proposed for extracting affine correspondences. We present an overview of the pipeline in Fig. \ref{overview}. Taking an image pair $I^A$, $I^B$ as input, the network produces reliable affine matches.

\subsection{Preliminary }

\noindent\hspace{1em} Let us assume that we are given a pair of images $I^A$ and $I^B$ and corresponding patches $patch^A$ and $patch^B$ in the images. 
Assuming that the objects captured in these patches are flat surfaces, there exists a linear transformation $\mathbf{A}$  satisfying $patch^A = \mathbf{A}$ $patch^B$, where $\mathbf{A} \in \mathbb{R}^{2\times2}$ is usually called a local affine transformation, described as $(a_{11}, a_{12}, a_{21}, a_{22})$, where $a_{ij}$ are the elements in a row-major order ($i,j \in{\{1, 2 \}}$). 

There are multiple ways to decompose an affine transformation. In this paper, we utilize the decomposition proposed in~\cite{affnet}, which decomposes to scale, orientation, and affine shape matrix $\mathbf{A}^\prime$. $\mathbf{A}^\prime$ is the affine shape matrix with det($\mathbf{A}^\prime$) = 1, which could be decomposed into identity matrix $\mathbf{I}$ and the residual shape $\mathbf{A}^{\prime\prime}$. The parameterizations of the affine transformation have a significant impact on the performance of local geometric estimators, as shown in~\cite{affnet}. 
Suppose that an affine correspondence ($\mathbf{p_1}$, $\mathbf{p_2}$, $\mathbf{A}$) and fundamental matrix $\mathbf{F}$ is known. It is trivial that every affine transformation preserves the direction of the lines going through points $\mathbf{p_1}$ and $\mathbf{p_2}$ in the first and second images~\cite{barath_toth_hajder_2017, 2ac2018barath}, where $\mathbf{p}_1 =\left[x_1, y_1, 1\right]^T$ and $\mathbf{p}_2$ $=\left[x_2, y_2, 1\right]^T$ represent a homogeneous form of point correspondence.
	
The geometric relationship of $\mathbf{p_1}$, $\mathbf{p_2}$, $\mathbf{F}$, and $\mathbf{A}$ is as:
\begin{equation}
\label{affine_con}
    \left( \mathbf{F} ^T \mathbf{p}_1\right)_{(1: 2)} +\left( \mathbf{A} ^T \mathbf{F} \mathbf{p} _2\right)_{(1: 2)} =0.
\end{equation}

\noindent\hspace{1em} The above equation only holds for the upper two rows and provides two linear equations.

\subsection{Affine Correspondence Estimation }

\noindent\hspace{1em} Each affine correspondence consists of a point correspondence and a local affine transformation. To improve the accuracy, we design a separate two-stage framework for optimization. Our overall process is first to extract the point matches and then estimate the affine transformation of their local patches. Unlike the traditional sparse feature matching method, we extract a large number of accurate keypoints from a dense warp and add a loss function with epipolar constraints to enable the network to learn more geometric information. 
The first sub-network is responsible for acquiring point correspondences and is employed to learn dense matching. Then, we obtain the corresponding patches according to the corresponding points. 
The second sub-network runs to estimate accurate local affine transformations, which are calculated through the scale, orientation, and residual shape. 
The final affine correspondences consist of these point matches and their affine transformations.

\subsubsection{Feature Matching Module}
\noindent\hspace{1em} In the first sub-network, we consider the task of estimating the accurate point correspondences from two images $\left(I^{ A }, I^{ B }\right)$. Inspired by DKM~\cite{DKM},  we choose the dense matching paradigm to estimate a dense warp $W ^{ A \rightarrow B }$ and a dense certainty $p^{ A \rightarrow B }$, which represents a dense warp and the probability of correct matches, respectively. Feature maps are extracted from $I^{ A }$ and $ I^{ B }$ with a ResNet50~\cite{2016Deep} encoder, which is pre-trained on ImageNet-1K~\cite{imagenet}. The initial encoding process could be described as
\begin{equation}
    \left(\Phi_{coarse}^{ A },\Phi_{fine}^{ A }\right)=E_\theta\left(I^{ A }\right),
\end{equation}
\begin{equation}
    \left(\Phi_{coarse}^{ B },\Phi_{fine}^{ B }\right)=E_\theta\left(I^{ B }\right),
\end{equation}
where $\Phi_{\text {coarse }}^A$, $\Phi_{\text {fine }}^A$, $\Phi_{\text {coarse }}^B$, and $\Phi_{\text {fine }}^B$  are feature maps obtained at different network depths. Function $E_\theta$ represents the encoder. After obtaining feature maps, a suitable regression framework is used to infer the mapping relationship of pixels. Through a global matcher, we can obtain the coarse-level warp and certainty, which is written as follows:	
%\begin{small}
\begin{equation}
    \left(\hat{ W }_{\text {coarse }}^{ A \rightarrow B }, \hat{p}_{\text {coarse }}^{ A \rightarrow B }\right)=G_\theta\left(\Phi_{\text {coarse }}^{ A }, \Phi_{\text {coarse }}^{ B }\right),
\end{equation}
%\end{small}
\noindent where $G_\theta$ is the kernel regression global matcher, which generates robust coarse matches, using an embedded Gaussian process regression. Parameter $\hat{W}_{\text {coarse }}^{A \rightarrow B}$ and $\hat{p}_{\text {coarse }}^{A \rightarrow B}$ are the coarse-level warp and certainty.

Meanwhile, to utilize the global features of the image, cosine encoding is used to enhance the ability to match weak textures and repetitive structures. The refining process predicts a residual discrepancy for the projected warp and a logit discrepancy for certainty. This process is reiterated until the highest resolution is attained as follows:	
%
%\begin{small}
\begin{equation}
    {
        \left(\hat{ W }_{\text {fine }}^{ A \rightarrow B }, \hat{p}_{\text {fine }}^{ A \rightarrow B }\right)=R_\theta\left(\Phi_{fine}^{ A }, \Phi_{fine}^{ B }, \hat{ W }_{\text {coarse }}^{ A \rightarrow B }, \hat{p}_{\text {coarse }}^{ A \rightarrow B }\right),
    }
\end{equation}
%\end{small}
\noindent where $\hat{ W }_{\text {fine}}^{ A \rightarrow B }$ and $\hat{p}_{\text {fine}}^{ A \rightarrow B }$ are the predicted warp and certainty in fine-level. Function $R_\theta$ is a set of refiners that uses stacked feature maps and depthwise convolution kernels.
Finally, the point correspondences with their patches are obtained as follows:
%
%\begin{small}
\begin{equation}
    \left({Patch}^{A }_l, {Patch}^{ B}_l\right)=S_\theta\left(\hat{ W }_{\text {fine }}^{ A \rightarrow B }, \hat{p}_{\text {fine }}^{ A \rightarrow B }\right),
\end{equation}
%\end{small}
\noindent where $\text{Patch}_l^A$ and $\text{Patch}_l^B$ are the corresponding patches cropped from $\left(I^{ A }, I^{ B }\right)$ with the size \emph{l}. Function $S_\theta$ is a sampler, selecting the matches. We obtain point correspondences by using warp and probability on each pixel. %, a better result than sparse matching. 
The final patch pairs are generated around these corresponding points, with a fixed patch size of 32*32 pixels.

\begin{figure*}[ht]
    \setlength{\abovedisplayskip}{3pt}
    \setlength{\belowdisplayskip}{3pt}
    \includegraphics[width=0.95\textwidth]{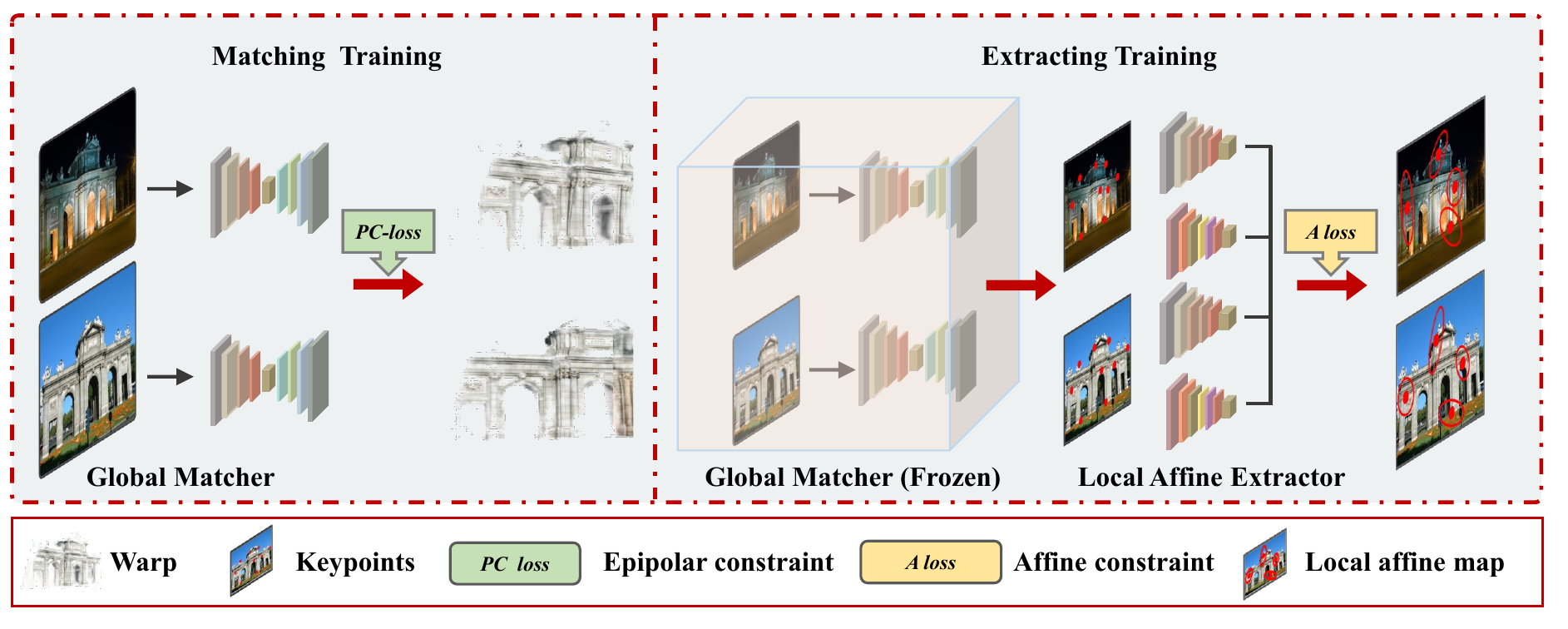}
    \caption{ \textbf{The training pipeline.} The network starts with training only the dense point matcher supervised by the proposed Sampson distance-based point correspondence loss.
    This network extracts a dense warp between the images.
    Next, the point matcher sub-network is frozen, and we train the affine shape extractor to minimize the proposed affine loss, leveraging the epipolar geometry-based constraints.
    }
    \label{decoupling}
\end{figure*}
 
\subsubsection{Local Affine Transformation Estimation Module}

\noindent\hspace{1em} After obtaining point correspondences and patches centered on these points using the above module, local affine transformations for these regions are estimated. The affine transformation is decomposed according to \cite{affnet}. The process of obtaining angles and scales is described as follows: 	
\begin{equation}
    \left({O}_{i}^A, {S}_{i}^A\right)=E_{o,s}\left(patch^A_l\right),
\end{equation}
\begin{equation}
    \left({O}_{i}^B, {S}_{i}^B\right)=E_{o,s}\left(patch^B_l\right),
\end{equation}
\begin{equation}
    O_i^{A \rightarrow B}={O}_{i}^B - {O}_{i}^A, \quad S_i^{A \rightarrow B}={S}_{i}^B/{S}_{i}^A,
\end{equation}
	
\noindent where $O_i^A$, $S_i^A$, $O_i^B$, and $S_i^B$ represent the orientations and scales of patch pairs. $E_{o,s}$ consists of two independent fully connected networks, by which we compute the relative scale and orientation of each patch correspondence. By discretizing angles and scales, this network can predict the probability distribution of patches at discrete angles and scales. The scale and angle with the highest probability are considered the predicted results.
The extraction of angles and scales is based on the SoTA method~\cite{S3Esti}. Through probabilistic covariant loss, the prediction accuracy of scale and direction is higher than that of traditional methods.
	
The final part is to calculate the residual shape, which is described as follows:
\begin{equation}
    \mathbf{A}_{i}^{\prime \prime }=E_{\text {aff }}\left(patch^A_l,patch^B_l\right),
\end{equation}
\noindent where the residual shape $\mathbf{A}^{\prime \prime}_i$, as described in \cite{affnet}, is computed via an independent fully connected network $E_{\text{aff}}$ used to regress the final residual shape. Finally, the affine correspondences are computed as represented through
\begin{equation}
    ACs=\Psi_{syn}(P_i^A,P_i^B,O_i^{A \rightarrow B},S_i^{A \rightarrow B},{A}_{i}^{\prime \prime}),
\end{equation}
\noindent where $\Psi_{syn}$ is the process of synthesizing affine correspondences according to \cite{affnet} .

\subsection{Loss Function}
\label{section_loss}

\noindent\hspace{1em} The epipolar loss has been demonstrated to be an effective way to improve matching performance~\cite{caps_2020}. 
To facilitate accurate detection, we further improve it by using the Sampson distance and also a loss based on affine features.

%Furthermore.
%We upgrade it using Sampson distance and add  affine transformation constraints as loss function. 
	
For training the matching, we make sure that the point correspondences agree with the epipolar geometry, and use loss $L_{\text{pc}}$ that is as follows: 
\begin{equation}\label{22}
    L _{\text {pc }}\left(\hat{PC}^{ A \rightarrow B }\right)=\frac{1}{N} \sum_{i=1}^N  SD_P(E_{PC}^i ),
\end{equation}
\begin{equation}\label{23}
    E_{PC}={\mathbf{p}_{2}^T\mathbf{F}\mathbf{p}_{1}} ,
\end{equation}
\noindent where $\mathbf{p}_1=\left(x_1, y_1, 1\right)^T$, $\mathbf{p}_2=\left(x_2, y_2, 1\right)^T$, and the $\mathbf{F}$ represent the fundamental matrix,  $i$ is the index of the point correspondence, 
function \emph{SD} calculates the Sampson Distance,  $\hat{PC}^{ A \rightarrow B }$ is the point correspondences, 
\emph{N} is the number of the correspondences, and 
$SD_P(E_{PC})$ is the Sampson Distance calculated from the epipolar constraint. 
The derivation of the Sampson Distance metric for epipolar constraints is put in the supplementary material.

For training the matcher, the used loss is as follows:

\begin{equation}
\begin{aligned}
    L_{\text{m}} = \sum_{l=1}^L & L_{\text{warp}}\left(\hat{W}^{A\rightarrow B}\right) && + \lambda L_{\text{conf}}\left(\hat{p}^{A\rightarrow B}\right) \\
                                 & && + \gamma L_{\text{pc}}\left(\hat{PC}^{A\rightarrow B}\right)
\end{aligned}
\end{equation}

\noindent where $\lambda$ and $\gamma$ are the balancing terms and set to 0.01 empirically. Specifically, for the warp loss $L_{\text{warp}}$, we use the $L_2$ distance of the predicted $W_l^{ A \rightarrow B }$ and the ground truth warp $\hat{W}_l^{ A \rightarrow B}$ as follows:
\begin{equation}
    L _{\text {warp }}\left(\hat{W}^{ A \rightarrow B }\right)=\sum_{\text {grid }} p_l \odot\left\|W^{ A \rightarrow B }-\hat{W}^{ A \rightarrow B }\right\|_2.
\end{equation}
For confidence loss $L_{\text{conf}}$, we use the unweighted binary cross entropy between the prediction confidence $\hat{p}_l$ and the ground truth $p_l$ written as follows:
%
%\begin{small}
\begin{equation}
    L _{\text {conf }}\left(\hat{p}\right)=\sum_{\text {grid }} p \log \hat{p}+\left(1-p\right) \log \left(1-\hat{p}\right).
\end{equation}
%\end{small}

Global matchers can provide precise point correspondences, but they are insufficient to obtain accurate affine correspondences. To train the affine sub-network, we introduce the novel affine constraint loss, which is crucial for learning the correct affine shape. 
It is designed to quantify how well our predicted shape complies with the epipolar geometry. %affine transformation constraint. 
Minimizing the affine constraint loss enables the network to estimate local affine shapes accurately. The loss function is as follows:
%
%\begin{small}
\begin{equation}
\label{sd_e}
    \begin{split}
        L _{\text {aff }}\left(\hat{AC}^{ A \rightarrow B }\right)=-\frac{1}{N} \sum_{i=1}^N SD_A(E_{AC}^i),
    \end{split}
\end{equation}
%\end{small}
\noindent where $\hat{AC}^{A \rightarrow B}$ is the obtained affine correspondences in images, $SD_A(E_{AC})$ is the affine transformation constraint error represented by the affine Sampson Distance, where  $E_{AC}$ is defined by Eq.~\ref{affine_con}.  
More specifically, it is:
\begin{equation}\label{sd_eac}
    SD_A(E_{AC})_{(1: 2)} = SD_A(\mathbf A^{-T}\left(\mathbf F^T \mathbf p_2\right)_{(1: 2)}+\left(\mathbf F \mathbf p_1\right)_{(1: 2)}).
\end{equation}
\noindent\hspace{1em} We included the derivation of the affine Sampson Distance metric in the supplementary material.

The loss function for extracting local affine shapes is as follows:
%\begin{small}
\begin{equation}\label{loss_extract}
        L_{\text{ext}}= L _{\text {aff }}\left(\hat{AC}^{ A \rightarrow B }\right) +  L(P, Q)_\text{ori} + L(P, Q)_\text{sca},
\end{equation}
%\end{small}

\noindent where $L(P, Q)_\text{ori}$ and $L(P, Q)_\text{sca}$ are probability covariance loss in the orientation and scale~\cite{S3Esti}. We discretize continuous angles and scales to convert regression into classification. Image patches are obtained through random scaling and rotation, using the scaling factor and rotation angle as labels. The network predicts discrete distributions of scale and orientation, and the loss function is constructed based on their discrepancy from the ideal distribution.
For orientation and scale, the loss is defined as:
%\begin{small}
\begin{equation}
    L_\text{ori} = \sum_{i}^{N} P_\text{ori} \log Q_\text{ori}, \ \
    L_\text{sca} = \sum_{i}^{N} P_\text{sca} \log Q_\text{sca},
\end{equation}
%\end{small}
%
respectively, where $P_i$ is the true discrete scale or orientation distribution, and $Q_i$ is the predicted discrete distribution.
These loss functions guide learning to ensure accurate and consistent affine correspondences.

\subsection{Decoupled Training Pipeline}

\noindent\hspace{1em} We employ a decoupled approach for training, as shown in Fig. \ref{decoupling}. 
The two sub-networks are trained separately to reduce the loss ambiguity caused by weak supervision. Using different losses helps with convergence and performance.
	
During the first part of the training, only the first sub-network is optimized to learn accurate matches, and the affine transformation part is ignored. Training the network until the loss no longer decreases, we stop and freeze parameters. Then, the second sub-network is trained for local affine transformation extraction. 
This two-step training leads to faster convergence and increased final performance compared to training the two sub-networks together. 
In addition, decoupling training consumes less memory than joint training.  This decoupled training approach avoids the loss of ambiguity caused by weak supervision strategies and greatly helps improve the performance of our network.
	
\section{Experiments}
\label{Section: Experiments}
\noindent\hspace{1em} In this section, we demonstrate superior performance compared to previous affine correspondence estimation and image matching methods. 
We validate the accuracy of the extracted matches by comparing with mainstream image-matching methods on the HPatches~\cite{Hptches}. %and ExtremeView (EVD)~\cite{MISHKIN201581}. 
Additionally, we compare the proposed method on relative pose estimation on the KITTI~\cite{KITTI} and MegaDepth~\cite{Li_Snavely_2018} datasets with the state-of-the-art matchers. Finally, we conduct ablation studies to verify the effectiveness of proposed component.
	
\subsection{Model Implementation}
\noindent\textbf{Dataset.}
We utilize the MegaDepth dataset~\cite{Li_Snavely_2018} and ScanNet~\cite{ScanNet} for training, using the same training and test split as in baseline approaches~\cite{DKM,sun2021loftr}.
 
\noindent\textbf{Implementation  details.} 
We use the matching sub-network to extract point correspondences and then select the exact corresponding points for the patches, whose size is set to 32*32 pixels. We train the patches for the subsequent affine shape extraction network with different loss functions, as mentioned in Section \ref{section_loss}.
The model is trained on ScanNet (indoor) and MegaDepth (outdoor) datasets separately. The pre-trained model is used for weight initialization, and only the weights of the Refiner module are fine-tuned during training. We trained the model using PC-loss by randomly sampling 2k points. The AdamW optimizer with a weight-decay of $10^{-2}$ is used. Then, the first sub-network is frozen while training the second one from scratch. The loss function for extracting affine shapes is employed. We employ SGD optimization with an initial learning rate of 0.0001 and adopt a learning rate decay strategy.

\subsection{Image Matching  with Affine Correspondences}

\noindent\hspace{1em} As our first experiment, we evaluate our method on the widely used HPatches dataset~\cite{Hptches}. Consistent with the approach in D2-Net ~\cite{d2net,posfeat}, we exclude 8 high-resolution scenes, leaving 52 scenes with illumination variations and 56 scenes with viewpoint changes for evaluation.

\noindent\textbf{Evaluation protocol.} We follow the setup proposed in Posfeat~\cite{posfeat} and report the Mean Matching Accuracy (MMA)~\cite{MMA} under thresholds varying from 1 to 10 pixels. We use a weighted sum of MMA at different thresholds for overall evaluation~\cite{posfeat} as follows: 
%\begin{small}
\begin{equation}\label{mma}
    \text { MMAscore }=\frac{\sum_{ thr \in[1,10]}(2-0.1 \cdot thr ) \cdot \text { MMA } @ thr }{\sum_{ thr \in[1,10]}(2-0.1 \cdot thr )}.
\end{equation}
%\end{small}
%

\begin{figure}
    \centering
    \setlength{\abovedisplayskip}{3pt}
    \setlength{\belowdisplayskip}{3pt}
    \includegraphics[width=0.47\textwidth]{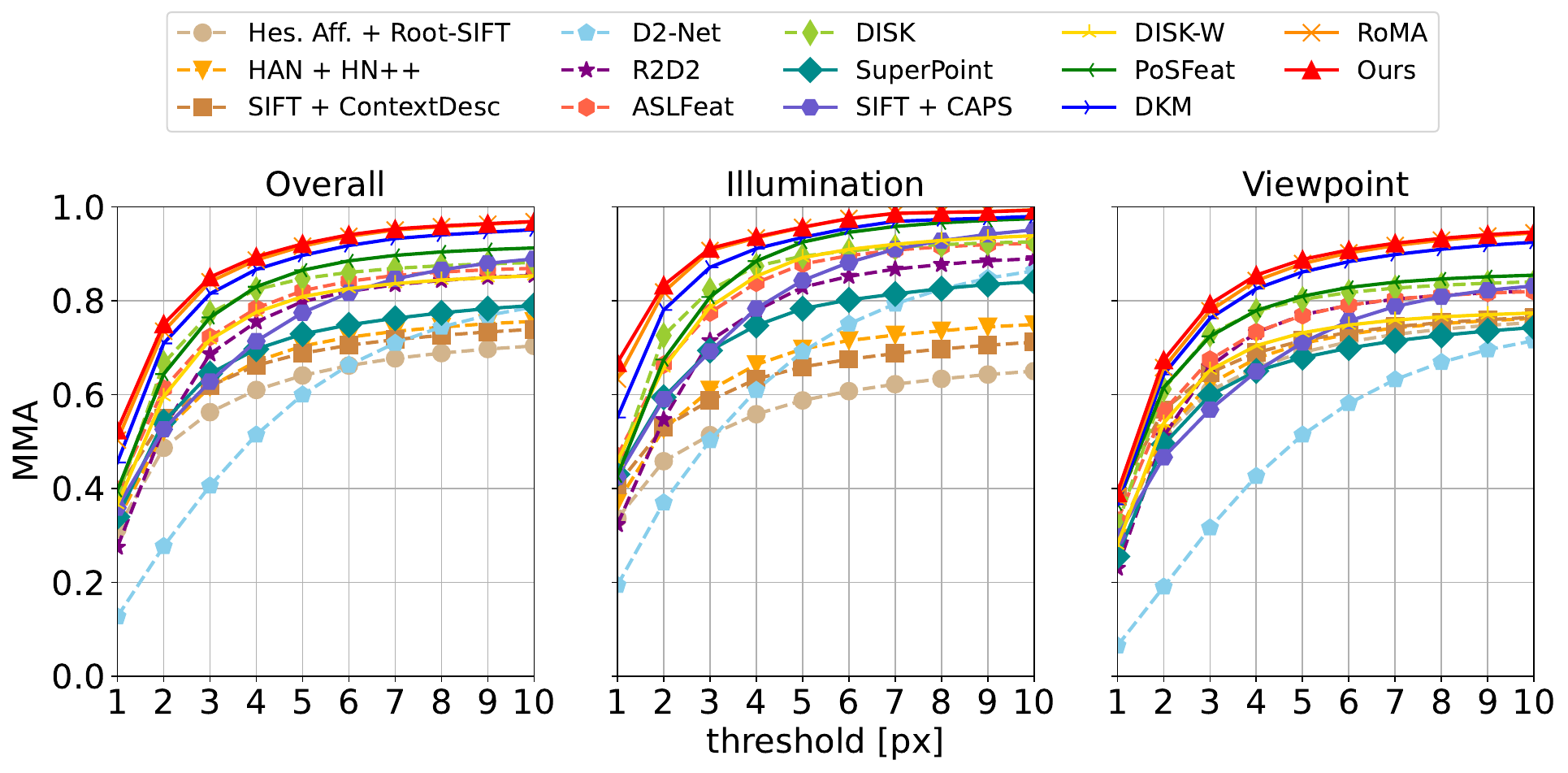}
    \caption{ The mean matching accuracy (MMA; higher is better) at different thresholds (in pixels) on the HPatches Dataset~\cite{Hptches}. }
    \label{fig:5}
\end{figure}

\begin{table}[t]
\caption{The weighted sum of mean matching accuracies at multiple thresholds (MMAscore, Eq.~\ref{mma}; higher is better) obtained by the baseline methods and the proposed one on the illumination and viewpoint sequences of the HPatches dataset~\cite{Hptches} separately, and on all. 
The best results are shown in bold.
}
\label{tab:1}
\centering
\huge
\renewcommand\arraystretch{1.1}
\resizebox{1\linewidth}{!}{
    \begin{tabular}{l|ccc}
        \hline \makebox[0.35\textwidth][c]{Methods} & \begin{tabular}{c} 
            MMAscore $\uparrow$ \\
            Overall
        \end{tabular} & \begin{tabular}{c} 
            MMAscore $\uparrow$ \\
            Illumination
        \end{tabular} & \begin{tabular}{c} 
            MMAscore $\uparrow$ \\
            Viewpoint
        \end{tabular} \\
        \hline 
        Hes. Aff.~\cite{2001hessian_affine} + Root-SIFT~\cite{haff} & 0.584 & 0.544 & 0.624 \\
        HAN~\cite{2001hessian_affine} + HN++~\cite{NIPS2017_831caa1b} & 0.633 & 0.634 & 0.633 \\
        SIFT~\cite{lowe_2004} + ContextDesc~\cite{luo_shen_zhou_zhang_yao_li_fang_quan_2019} & 0.636 & 0.613 & 0.657 \\
        D2Net~\cite{d2net} & 0.519 & 0.605 & 0.440 \\
        R2D2~\cite{r2d2} & 0.695 & 0.727 & 0.665 \\
        ASLFeat~\cite{luo2020aslfeat} & 0.739 & 0.795 & 0.687 \\
        DISK~\cite{disk} & 0.763 & 0.813 & $0.716$ \\
        DELF~\cite{delf} & 0.571 & $0.903$ & 0.262 \\
        SuperPoint~\cite{SuperPoint} & 0.658 & 0.715 & 0.606 \\
        SIFT~\cite{lowe_2004} + CAPS ~\cite{caps_2020} & 0.699 & 0.764 & 0.639 \\
        DISK-W~\cite{disk} & 0.719 & 0.803 & 0.649 \\
        PoSFeat~\cite{posfeat}  & 0.775 & $0.826$ & $0 . 7 2 8$ \\
        DKM~\cite{DKM}  & 0.819 & $0.869$ & $0.772$ \\
        RoMA~\cite{roma}  & 0.843 & $0.901$ &  $0.789$ \\
        \hline
        \textbf{Ours} & \textbf{0.851} & $\textbf{0.908}$ & $\textbf{0.798 }$ \\
        \hline
    \end{tabular}
    }
\end{table}
\noindent\textbf{Result.} 
As shown in Fig.~\ref{fig:5} and Table~\ref{tab:1}, the proposed method achieves the highest MMA scores on both the illumination and viewpoint sequences. Our method outperforms the previous SoTA~\cite{roma} and achieves substantially better overall performance of 0.851 MMAscore. This demonstrates that accurate local affine shapes can provide additional cues for matching that further improve accuracy.

\subsection{Improvement in the Affine Frames}
%\subsubsection{Affine Frames Comparison on the HPatches Dataset}

\noindent\hspace{1em} While in the previous experiment, we focused on showcasing the improved accuracy of the point correspondences, now we demonstrate the accuracy of the affine frames themselves on the HPatches dataset~\cite{Hptches}.
We extract ground truth affine shapes at each point location from the ground truth homography as proposed in~\cite{Barath_Hajder_2017}.

We compare the proposed method with the view-synthesis-based Affine-SIFT  (ASIFT)~\cite{ASIFT}, the VLFeat library~\cite{Vedaldi2010VlfeatAO}, and the learning-based AffNet~\cite{affnet}.
We evaluate the similarity of the ground truth affine matrix and the estimated one by the Euclidean distance and cosine similarity. 

Table \ref{A_result} shows that the estimated affine matrices exhibit higher cosine similarity and smaller Euclidean distance than the baselines compared to the ground truth affine shapes.
This demonstrates that the proposed method estimates not only accurate keypoints but also precise affine shapes.

\begin{table}[ht]
    \centering
    \caption{The accuracy of the affine shapes estimated by the VLFeat library~\cite{Vedaldi2010VlfeatAO}, ASIFT~\cite{ASIFT}, AffNet~\cite{affnet} and the proposed method on the HPatches dataset~\cite{Hptches}.
    Reported metrics (bold=best): Euclidean distance and cosine similarity of affine matrices vs. ground truth. 
    The best results are in bold.
    }
    \label{A_result}
    \resizebox{1\linewidth}{!}{
        \setlength{\tabcolsep}{2mm}{
            \begin{tabular}{l c c c c c}
                \hline
                \multirow[b]{2}{*} &VLFeat~\cite{Vedaldi2010VlfeatAO} & AffNet~\cite{affnet} & ASIFT~\cite{ASIFT}  & \textbf{Ours} \\
                \hline
                Euclidean-Distance $\downarrow$ & 0.202 & 0.264 &  0.329& \textbf{0.123} \\
                Cosine-Similarity $\uparrow$ &  0.988 & 0.973 &0.894 & \textbf{0.994} \\
                \hline
            \end{tabular}
        }
    }
\end{table}

\subsection{Relative Pose Estimation}

\noindent\hspace{1em} In the previous sections, we demonstrate that the proposed method extracts more accurate point and local affine transformation than the state-of-the-art approaches. 
Here, we demonstrate that the affine correspondences obtained by our method lead to improved relative pose accuracy compared with methods obtaining point correspondences. 
More experiments can be found in the supplementary materials. 

\begin{table}[t]
    \tiny
    \centering
    \caption{
    Relative pose Accuracy Under the recall Curve (AUC; higher is better) thresholded at $5^\circ$, $10^\circ$, and $20^\circ$ on the MegaDepth~\cite{Li_Snavely_2018} dataset. 
    All methods run RANSAC-based essential matrix estimation, except for the last row, where we run the affine-based GC-RANSAC~\cite{gc-ransac2022,barath_polic_förstner_sattler_pajdla_kukelova_2020}, benefiting from the affine correspondences that we obtain.
    The best results are in bold.
    }
    \label{tab:2}
    \renewcommand\arraystretch{1.2}
    \resizebox{1\linewidth}{!}{
    \begin{tabular}{l c c c }
        \hline Method \hspace{3em} AUC @ $\rightarrow$ & $5^{\circ} \uparrow$ & $10^{\circ} \uparrow$ & $20^{\circ} \uparrow$ \\
        \hline LoFTR~\cite{sun2021loftr} {\tiny$\text{CVPR} 21$ }& 52.8 & 69.2 & 81.2 \\
        \hline ASpanFormer~\cite{Chen2022ASpanFormerDI} {\tiny$\text{ECCV}{22}$} & 55.3 & 71.5 & 83.1 \\
        \hline PDC-Net+~\cite{PDC-Net+} {\tiny$\text{TPAMI}{23}$} & 51.5 & 67.2 & 78.5 \\
        % \hline LightGlue~\cite{LightGlue}{\tiny$ICCV ^{23}$ }& 51.0 & 68.1 & 80.7 \\
        \hline DKM~\cite{DKM} {\tiny$\text{CVPR}{23}$ }& 60.4 & 74.9 & 85.1 \\
        \hline ROMA~\cite{roma} {\tiny$\text{CVPR}{24}$ }& 62.6 & 76.7 & \textbf{86.3} \\
        \hline \textbf{Ours (RANSAC)} & 63.1 & 76.3 & 85.2 \\
        \hline \textbf{Ours (affine GC-RANSAC)}  & \textbf{65.5} & \textbf{77.9} & \textbf{86.3} \\
        \hline
    \end{tabular}
            }

\end{table}

\subsubsection{Relative Pose Estimation on MegaDepth-1500}

\noindent\hspace{1em} We use the  MegaDepth-1500 dataset that consists of 1500 pairs from scene 0015 (St. Peter’s Basilica) and 0022 (Brandenburg Gate)~\cite{Li_Snavely_2018}. 
We follow the evaluation protocol in~\cite{superglue,sun2021loftr} and use a RANSAC threshold of 0.5 pixel. 
	
\noindent\textbf{Evaluation protocol.} 
Following~\cite{superglue} and~\cite{sun2021loftr}, we report the AUC of the pose error at thresholds $\left(5^{\circ}, 10^{\circ}, 20^{\circ}\right.$).
To compare with existing methods on the same baseline, we utilize RANSAC as implemented in the OpenCV library as previous methods do~\cite{sun2021loftr}. 
To demonstrate that the estimated affine frames are beneficial for pose estimation, we also run the affine correspondence-based Graph-Cut RANSAC~\cite{gc-ransac2022,barath_polic_förstner_sattler_pajdla_kukelova_2020}, designed specifically to leverage affine shapes together with the point locations.
 
\noindent\textbf{Results.} 
As shown in Table \ref{tab:2}, the proposed method with RANSAC achieves similar results to the state-of-the-art RoMA~\cite{roma} matcher.
When leveraging the estimated affine shapes with GC-RANSAC, the proposed method achieves the best performance. 
It outperforms DKM~\cite{DKM} by a significant 5.1 AUC points at $5^{\circ}$, while also being better than the recent RoMA by 2.4 AUC points.
These results demonstrate that the proposed geometric constraints can significantly improve matching based on feature descriptors.

\begin{table*}[ht]
\caption{The rotation and translation RMSE of the estimated relative poses on sequences 00 to 10 of the KITTI~\cite{KITTI} dataset. 
    Point-based methods, DKM~\cite{DKM} and RoMA~\cite{roma}, run RANSAC-based essential matrix estimation.
    Affine-based methods, VLFeat~\cite{Vedaldi2010VlfeatAO}, AffNet~\cite{affnet}, ASIFT~\cite{ASIFT}, and the proposed one, run the affine-based GC-RANSAC~\cite{gc-ransac2022,barath_polic_förstner_sattler_pajdla_kukelova_2020}, directly benefiting from the affine correspondences.}
\label{tab_relative}
\renewcommand\arraystretch{1.1}
\resizebox{\linewidth}{!}{
    \begin{tabular}{l|c|c|c|c|c|c|c|c|c|c|c|c}
        \hline  & \multicolumn{6}{c|}{ Rotation RMSE ($^\circ$) $\downarrow$ } & \multicolumn{6}{c}{ Translation RMSE ($^\circ$) $\downarrow$ } \\
        \hline 
        Solver &
        VLFeat~\cite{Vedaldi2010VlfeatAO} &  AffNet~\cite{affnet}&ASIFT~\cite{ASIFT}  & DKM~\cite{DKM} & RoMA~\cite{roma}& \textbf{Ours} &   VLFeat~\cite{Vedaldi2010VlfeatAO} & AffNet~\cite{affnet} & ASIFT~\cite{ASIFT}& DKM~\cite{DKM}& RoMA~\cite{roma}  & \textbf{Ours} \\
        \hline 
        Seq. 0 & 0.0467 & 0.0625 & 0.0360& 0.0429 & 0.0406 &\textbf{0.0353} & 0.799 & 0.697 & 0.968 & 0.751 & 0.735 &\textbf{0.689} \\
        Seq. 1 & 0.0433 & 0.0297 & 0.0366& 0.0428 & 0.0399 &\textbf{0.0287} & 0.690 & 0.657 & 0.648 & 0.603 & \textbf{0.590} &0.603 \\
        Seq. 2 & 0.0395 & 0.0332 & 0.0627& 0.0389 & 0.0375 &\textbf{0.0320} & 0.732 & 0.678 & 0.979 & 0.739 & 0.726 &\textbf{0.673} \\
        Seq. 3 & 0.0434 & 0.0390 & 0.0766& 0.0427 & 0.0409 &\textbf{0.0378} & 0.665 & 0.641 & 0.585 & 0.655 & 0.615 &\textbf{0.574} \\
        Seq. 4 & 0.0285 & 0.0213 & 0.0529& 0.0278 & 0.0279 &\textbf{0.0200} & 0.389 & 0.420 & 0.894 & 0.422 & 0.398 &\textbf{0.350} \\
        Seq. 5 & 0.0567 & 0.0277 & 0.1100& 0.0335 & 0.0312 &\textbf{0.0263} & 0.737 & 0.463 & 1.379 & 0.471 & 0.450 &\textbf{0.407} \\
        Seq. 6 & 0.0447 & 0.0227 & 0.0640& 0.0290 & 0.0272 &\textbf{0.0214} & 0.486 & 0.362 & 0.715 & 0.376 & 0.371 &\textbf{0.353} \\
        Seq. 7 & 0.0397 & 0.0269 & 0.0650& 0.0311 & 0.0295 &\textbf{0.0262} & 0.780 & 0.638 & 1.190 & 0.676 & 0.659 &\textbf{0.610} \\
        Seq. 8 & 0.0366 & 0.0281 & 0.0557& 0.0349 & 0.0325 &\textbf{0.0271} & 0.937 & 0.861 & 1.120 & 0.899 & 0.875 &\textbf{0.842} \\
        Seq. 9 & 0.0369 & 0.0295 & 0.0516& 0.0353 & 0.0329 &\textbf{0.0279} & 0.508 & 0.464 & 0.619 & 0.489 & 0.483 &\textbf{0.439} \\
        Seq. 10 & 0.0558& 0.0383 & 0.1090& 0.0394 & 0.0369 &\textbf{0.0348} & 0.733 & 0.599 & 1.180 & 0.558 & 0.564 &\textbf{0.543} \\
        \hline
        Average & 0.0429 & 0.0326& 0.0655& 0.0362& 0.0343 & \textbf{0.0289} & 0.678 & 0.587& 0.934 & 0.604& 0.587& \textbf{0.554} \\
        \hline
    \end{tabular}
}
\end{table*}

\subsubsection{Relative Pose Estimation on KITTI}

\noindent\hspace{1em} To further verify the accuracy of the affine correspondences, we apply the method to each consecutive pair of stereo pairs in the 11 test KITTI sequences~\cite{motion_from_two_AC}. 
The proposed method is compared with the most widely used affine correspondences extraction methods, including ASIFT~\cite{ASIFT}, AffNet~\cite{affnet}, and VLFeat~\cite{Vedaldi2010VlfeatAO,barath_polic_förstner_sattler_pajdla_kukelova_2020}. 
Also, we include the results of point-based methods, like DKM~\cite{DKM} and RoMA~\cite{roma}.
Similarly as on the MegaDepth dataset, affine-based methods use the affine GC-RANSAC method, while point-based ones use RANSAC.

\noindent\textbf{Evaluation protocol}. 
Rotation and translation errors are measured using RMSE. Rotation error is the angular difference between the ground truth and estimated rotation, while translation error is evaluated similarly by comparing angular differences, as done in baseline methods. %Our definition is as follows~\cite{guan_eccv_2022}.
	
\noindent\textbf{Result}. The performance is evaluated based on the median error for each image sequence. Table \ref{tab_relative} presents the per-frame error in rotation and translation direction for all tested KITTI sequences, calculated according to the provided ground truth.
The proposed method improves the rotation accuracy in all test sequences.
We also improve the estimated translations in all but one sequence, where the proposed method secures the second lowest errors. 

\subsection{Ablation Study}
\noindent\hspace{1em} To validate the effectiveness of designed components, we conduct ablation experiments on the HPatches~\cite{Hptches} and MegaDepth~\cite{Li_Snavely_2018} datasets.

First, We first compared the performance of different affine 558
transformation synthesis methods. we compare the results from direct affine shape regression, approximating the affine shape
from orientation and scale, and, finally, using direction, scale
and residual shape to estimate the affine transformation. 
As shown in Table \ref{Calculate_A}, among the several ways to calculate the affine transformation matrix, the way we adopt leads to the highest degree of similarity to the ground truth. This is consistent with the conclusion in AffNet~\cite{affnet}.
Second, we demonstrate the improvement caused by our proposed PC-loss, affine extractor, and affine loss when added to DKM~\cite{DKM} on relative pose estimation on the MegaDepth dataset~\cite{Li_Snavely_2018}. 
Table~\ref{tab:ablation_megadepth} demonstrates that each proposed component leads to improvements in all accuracy metrics compared to the original DKM method.

\begin{table}[ht]
    \centering
    \caption{Ablation study on affine shape parameterizations on the HPatches~\cite{Hptches} dataset. 
    The reported metrics are the Euclidean distance and cosine similarity of the estimated affine matrices w.r.t.\ the ground truth ones.
    n/c – did not converge.}
    \label{Calculate_A}
    \resizebox{0.95\linewidth}{!}{
    \begin{tabular}{ l c c c}
       \hline Estimated parameters  & Euclidean-Distance & Cosine-Similarity \\
       \hline  $\left(a_{11}, a_{12}, a_{21}, a_{22}\right)$  & n/c & n/c \\
       \hline $\left(O,S\right)$  & 0.346 & 0.987 \\
       \hline  $\left(O,S,A^{\prime \prime}\right)$  & $\textbf{0.123}$ & $\textbf{0.994}$ \\
       \hline
    \end{tabular}
}
\end{table}

\begin{table}[ht]
    \centering
    \caption{Ablation study on MegaDepth-1500~\cite{Li_Snavely_2018}. We report the relative pose Accuracy Under the recall Curve (AUC; higher is better) thresholded at $5^\circ$, $10^\circ$, and $20^\circ$. 
    The best results are in bold. }
    \label{tab:4}
    \resizebox{0.95\linewidth}{!}
    {\begin{tabular}{l c c c }
        \hline Method \hspace{10em} AUC @ $\rightarrow$ & $5^{\circ} \uparrow$ & $10^{\circ} \uparrow$ & $20^{\circ} \uparrow$ \\
        \hline DKM  & 60.4 & 74.9 & 85.1 \\
        \hline DKM + PC-loss & 60.7 & 75.1 & \textbf{85.4} \\
        \hline \textbf{Ours (DKM  + Affine Extractor + PC-Loss + AC-Loss)}  & \textbf{63.1} & \textbf{76.3} & 85.2 \\
        \hline
    \end{tabular} }
    \label{tab:ablation_megadepth}
    \vspace{-0.1cm}
\end{table}

\section{Conclusion}

\label{Section: Conclusion}

\noindent\hspace{1em} We propose a new framework designed for affine correspondence extraction. The geometric constraints that the points and affine shapes induce are formalized as losses used to supervise the network, learning geometry to improve matching accuracy. 
The experiments demonstrate that the proposed method surpasses existing affine shape estimators in terms of accuracy, while also improving upon state-of-the-art point-based approaches. 
We believe this research advances accurate affine correspondence extraction.

\noindent\textbf{Acknowledgments}
{\small
\noindent\hspace{1em} This work was supported by the National Natural Science Foundation of China (Grant No. 12372189) and the Hunan Provincial Natural Science Foundation for Excellent Young Scholars (Grant No. 2023JJ20045).
}

{
    \small
    \bibliographystyle{ieeenat_fullname}
    \bibliography{sec/11_references}
}

% \input{sec/12_appendix}

\end{document}

\title{ Supplementary Material – Learning Affine Correspondences by Integrating Geometric Constraints}

\author{
Pengju Sun\textsuperscript{1, 2}\hspace{0.2em}  Banglei Guan\textsuperscript{1, 2(\Letter)}\hspace{0.2em}  Zhenbao Yu\textsuperscript{1, 2} \hspace{0.2em} Yang Shang\textsuperscript{1, 2}\hspace{0.2em}  Qifeng Yu\textsuperscript{1, 2}\hspace{0.2em}  Daniel Barath\textsuperscript{3, 4}\\[2mm]
	{$^{1}$College of Aerospace Science and Engineering,}  { National University of Defense Technology, China.} \\[0.5mm]
    {$^{2}$   Hunan Provincial Key Laboratory of Image Measurement and Vision Navigation, China.} \\[0.5mm]
	{$^{3}$ETH Zurich, Switzerland.}
    {$^{4}$ HUN-REN SZTAKI, Hungary.}
}

\maketitle

% \maketitlesupplementary

\section{Overview}
\label{sec:overview}
\noindent\hspace{1em} In this supplementary material, we provide the details of the loss function and additional experiments. In Sec.~\ref{sec:loss function}, the geometric constraints based on Sampson Distance are derived. In Sec.~\ref{Image-Matching on EVD}, we demonstrate the performance of our method on datasets with large viewpoint changes. 
In Sec.~\ref{ScanNet-1500}, we demonstrate that the affine correspondences obtained by our method lead to improved relative pose accuracy compared with methods obtaining point correspondences on the indoor dataset. In Sec.~\ref{sec:Failed Cases}, we show the failed cases.

\section{Details of Sampson Distance for Geometric Constraints}
\label{sec:loss function}
\noindent\hspace{1em} Sampson Distance was originally introduced for conic fitting. 
The method finds the refined parameters that reduce
the overall fitting errors iteratively~\cite{Sampson_Distance}. 
Recently, Sampson Distance has also been used to model the measurement residuals of the correspondences between two views in computer vision~\cite {Dai2016RollingSC}. It can be regarded as a first-order approximation of geometric error and offers an efficient and effective alternative to traditional error metrics. Characterized by its reduced computational complexity, it provides an estimate of error that is comparable in accuracy to the geometrical error~\cite{Sampson_Distance}. 
In previous work, Zhou et al.~\cite{Patch2Pix} proposed that how much a match prediction fulfills the epipolar geometry can be precisely measured by the Sampson distance. 
In this paper, A novel affine transformation loss, represented by the Affine Sampson Distance, is introduced to further enhance the conformity of affine correspondences with the scene geometry. 
Given an AC satisfying $G_E( \widehat{ X } )=0$, where $G_E( X )$ is the geometric constraint  approximated by a Taylor expansion:

\begin{equation}
\label{reb_eq1}
G_E\left( X + \delta _{ X }\right) \approx G_E( X )+\frac{\partial G_E}{\partial X } \delta _{ X },
\end{equation}
$\delta _{ X }$ quantifies the measurement residual. Letting  
\begin{equation}
J = \frac{\partial G_E}{\partial X},
\end{equation}

\begin{equation}
\epsilon = G_E(X) - G_E(\widehat{X}), 
\end{equation} 
namely,

\begin{equation}
 J \delta_X = -\epsilon ,
 \end{equation}
 
the goal is to find \( \delta_X \) that minimizes \( \|\delta_X\| \) subject to Eq.~\ref{reb_eq1}. The problem can be solved by Lagrange Multipliers and the Sampson Distance is defined as the squared norm of $\delta _{ X }$.    

\begin{equation}
\left\| \delta _{ X }\right\|^2= \epsilon ^T\left(J J^T\right)^{-1} \epsilon.
\end{equation}

\noindent For the epipolar constraints,
\begin{equation}
G_E(X)=p_2^T F p_1.
\end{equation}

We take partial derivatives of $x_1$, $y_1$, $x_2$, $y_2$. Let $Z_0 = G(X)$. The remaining terms are $Z_1=\frac{\partial Z_0}{\partial x_1}$, $Z_2=\frac{\partial Z_0}{\partial y_1 }$, $Z_3=\frac{\partial Z_0}{\partial x_2 }$, $Z_4=\frac{\partial Z_0}{\partial y_2}$, 
we can obtain Eq.~\ref{sdF_pc2}.
 
\begin{figure*}    
    \centering
    \subfloat[ASIFT]{
        \begin{minipage}[b]{0.2\linewidth}
            \centering
            \includegraphics[width=\linewidth]{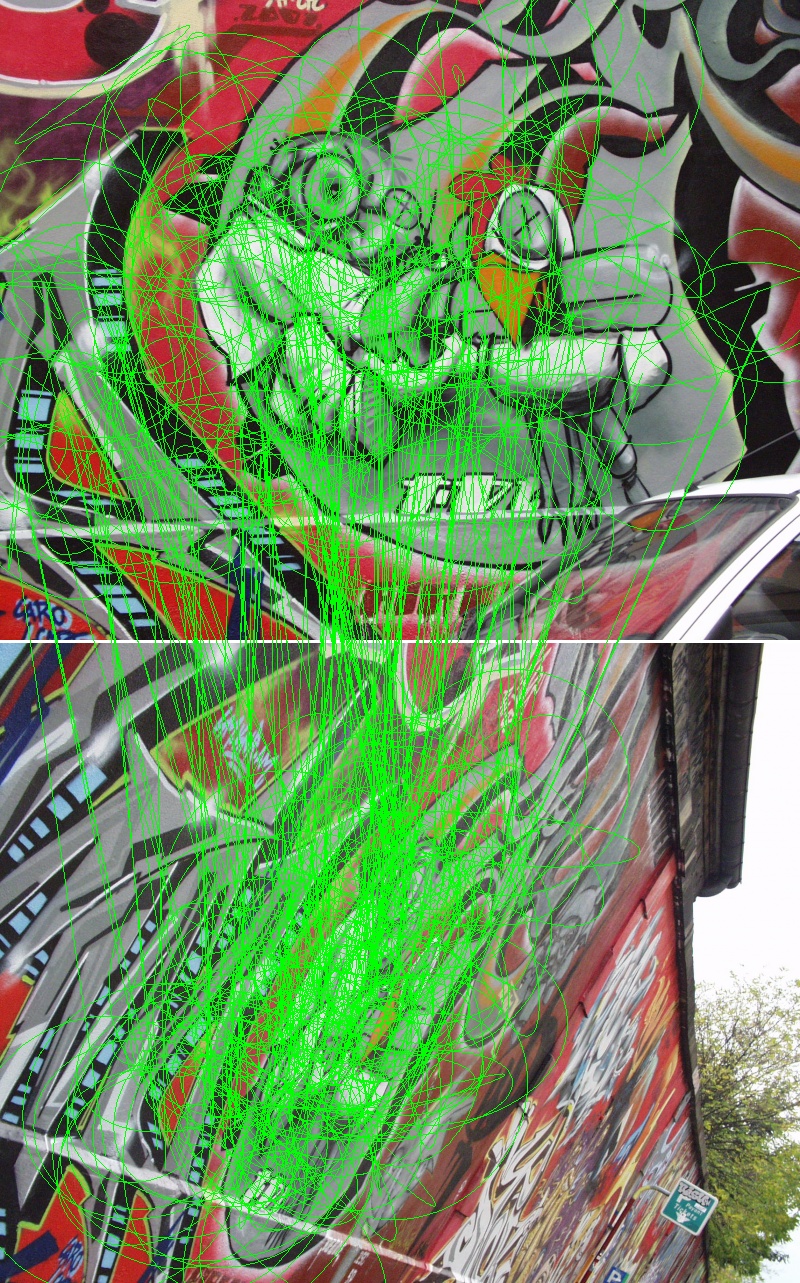}\\
            \vspace{0.2cm}
            \includegraphics[width=\linewidth]{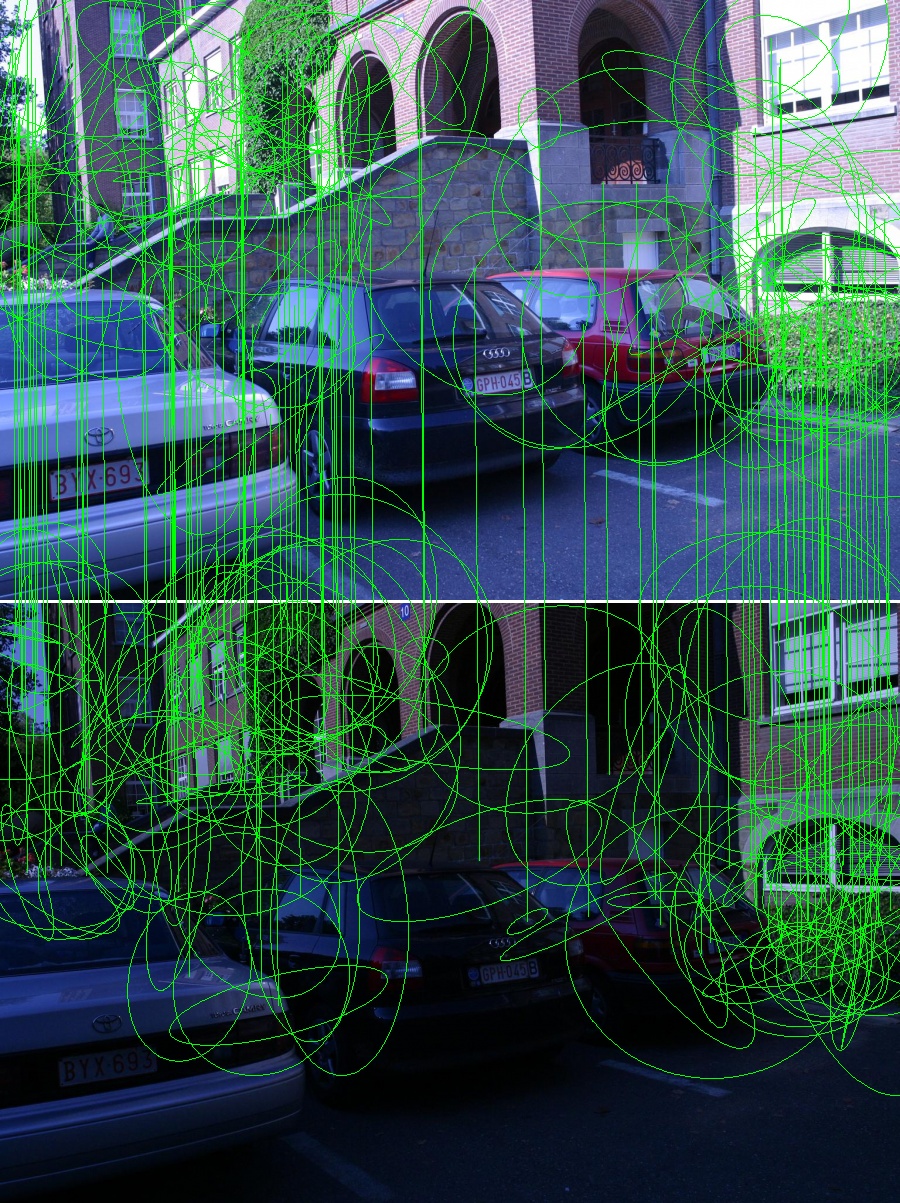}\\
            \vspace{0.2cm}
            \includegraphics[width=\linewidth]{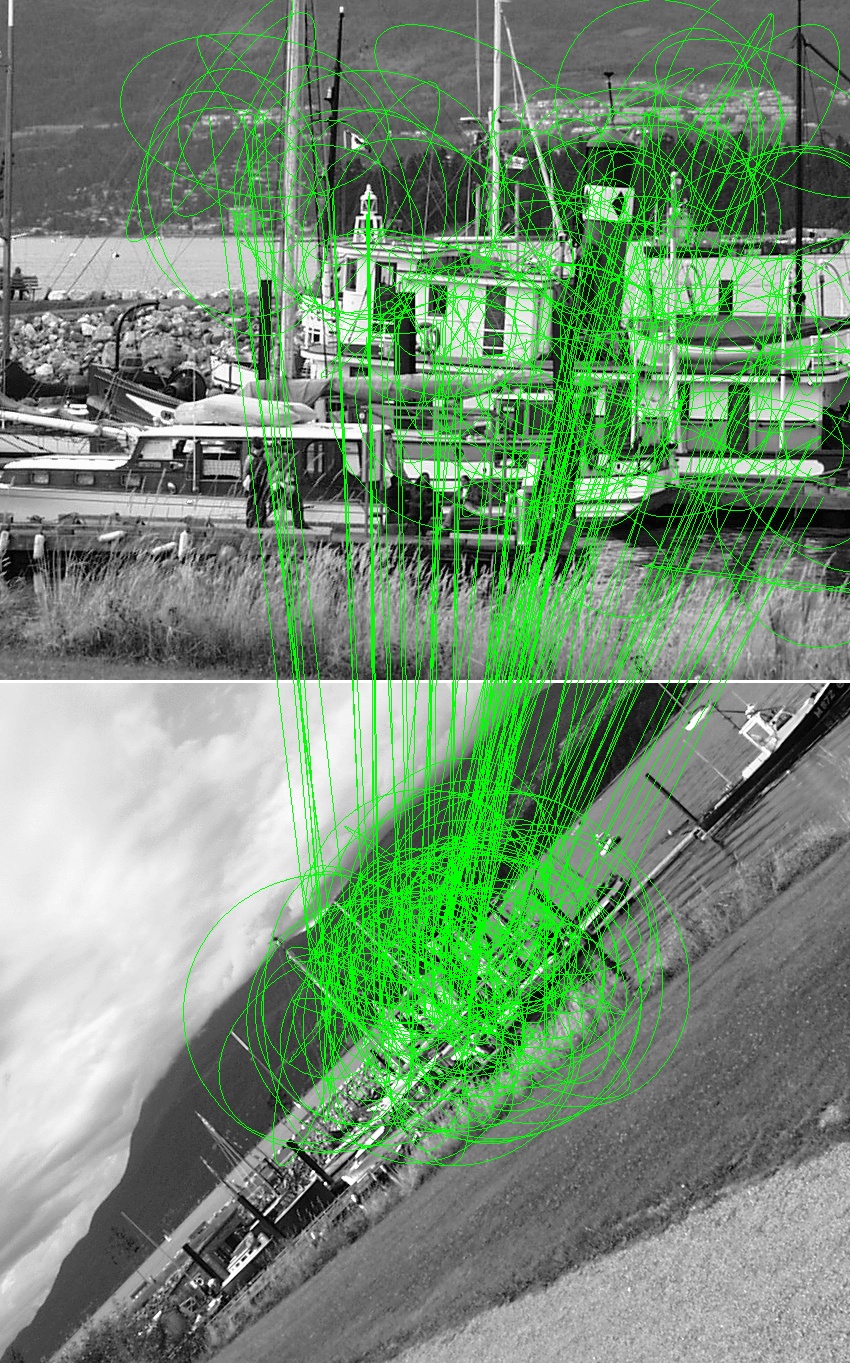}\\
        \end{minipage}
    }
    \vspace{0.02cm}
    \subfloat[VLFeat]{
        \begin{minipage}[b]{0.2\linewidth}
            \centering
            \includegraphics[width=\linewidth]{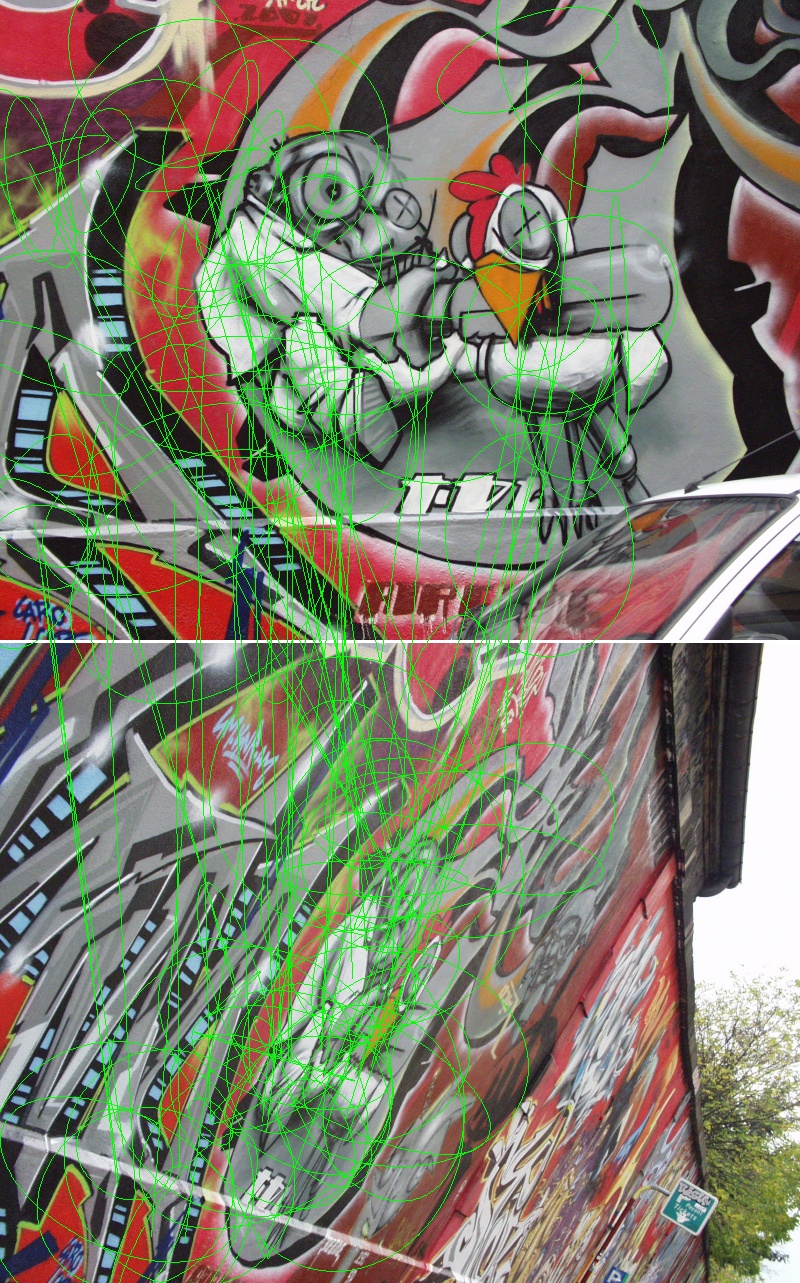}\\
            \vspace{0.2cm}
            \includegraphics[width=\linewidth]{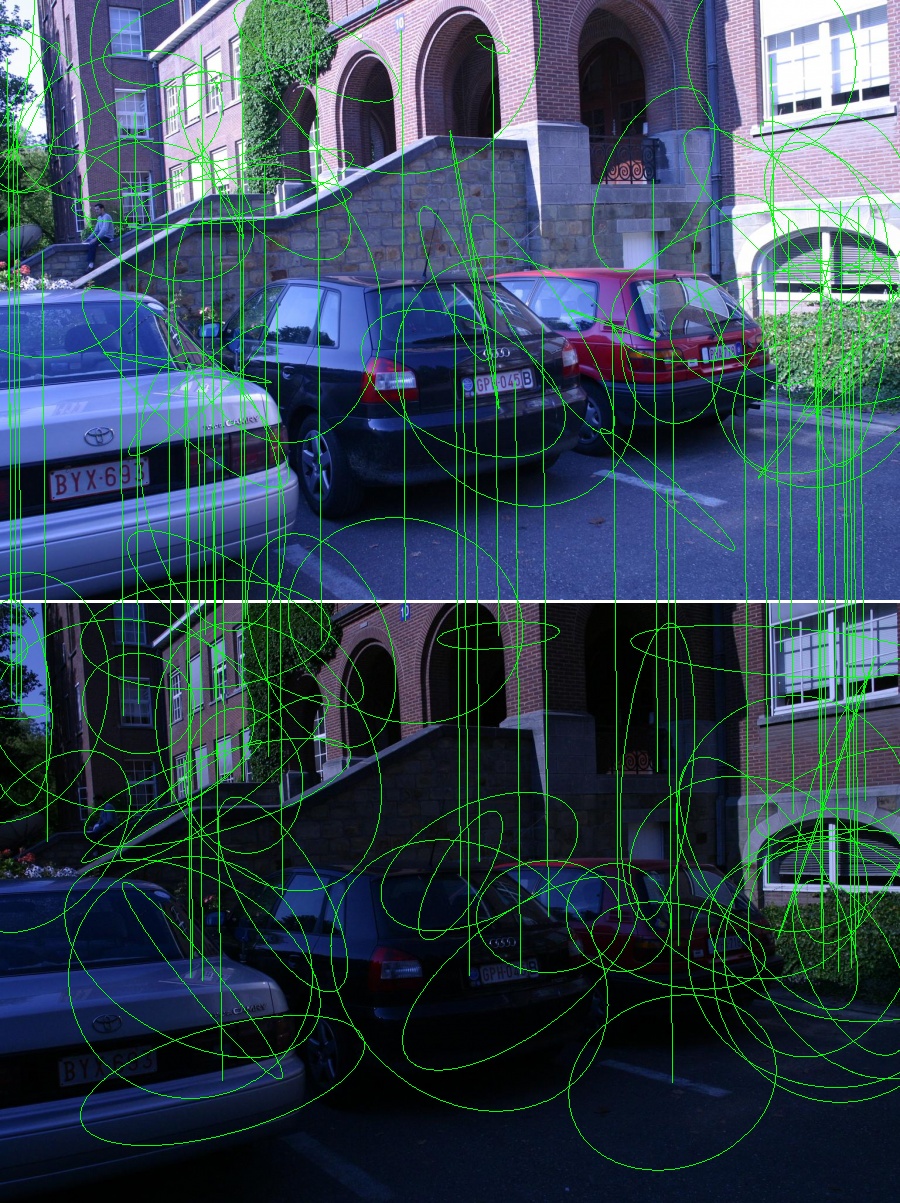}\\
            \vspace{0.2cm}
            \includegraphics[width=\linewidth]{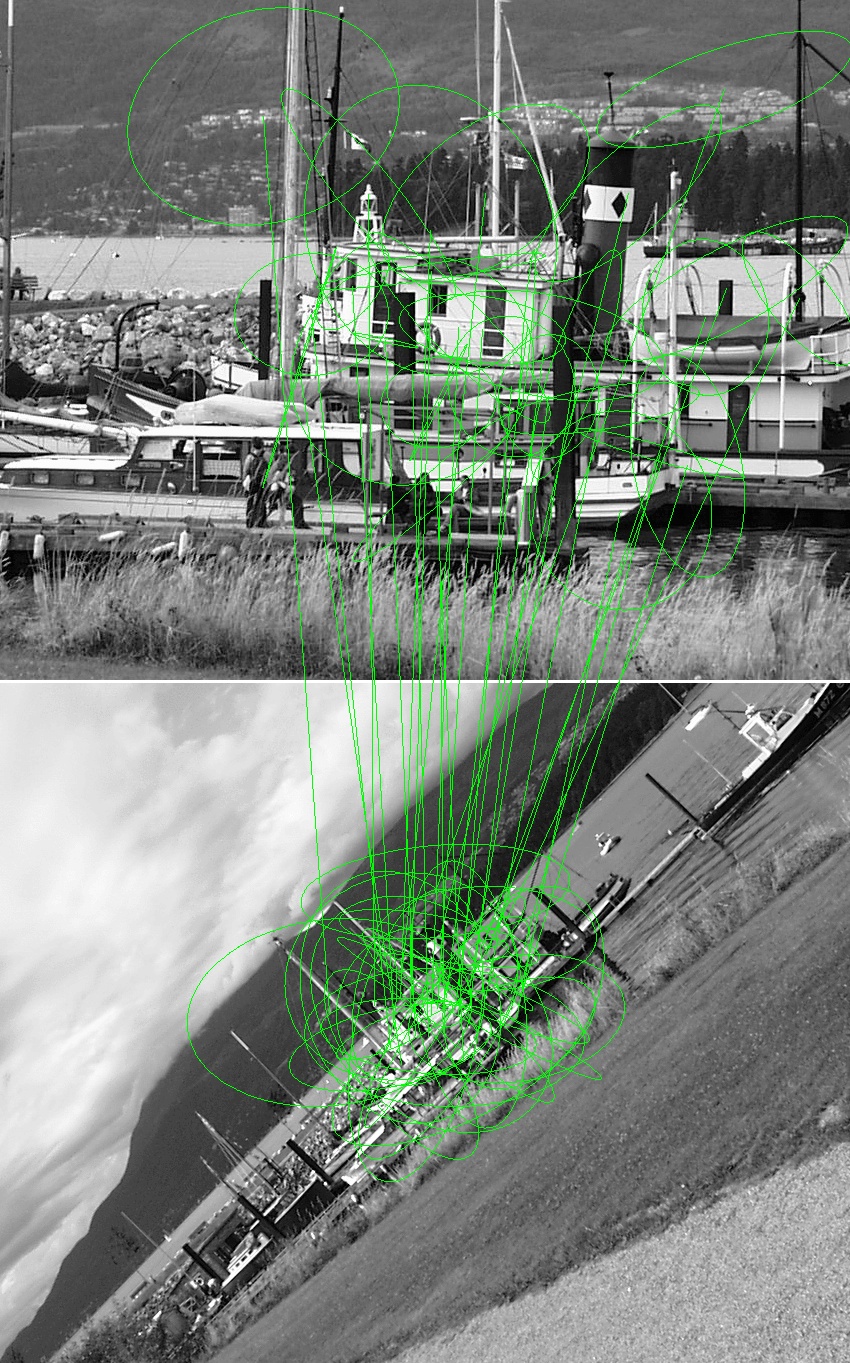}\\       
        \end{minipage}
    }
    \vspace{0.02cm}
    \subfloat[AffNet]{
        \begin{minipage}[b]{0.2\linewidth}
            \centering
            \includegraphics[width=\linewidth]{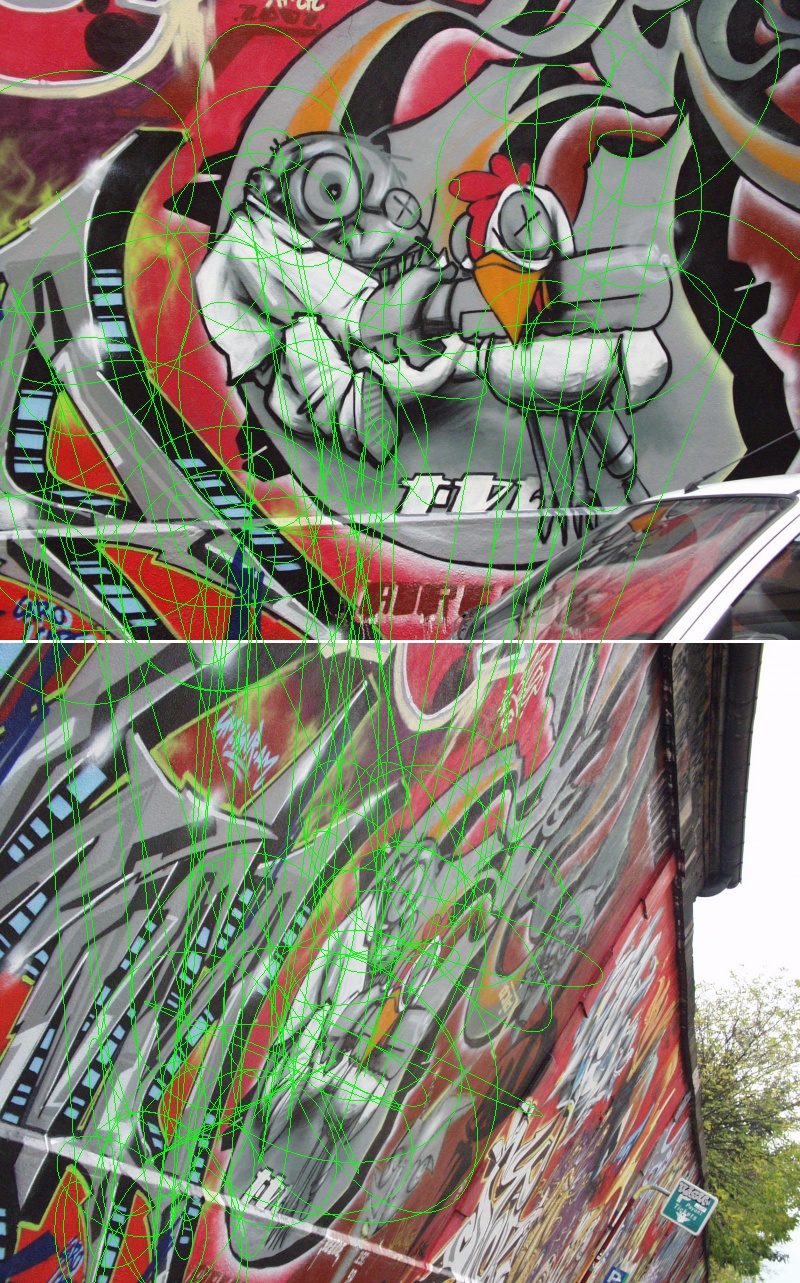}\\
            \vspace{0.2cm}
            \includegraphics[width=\linewidth]{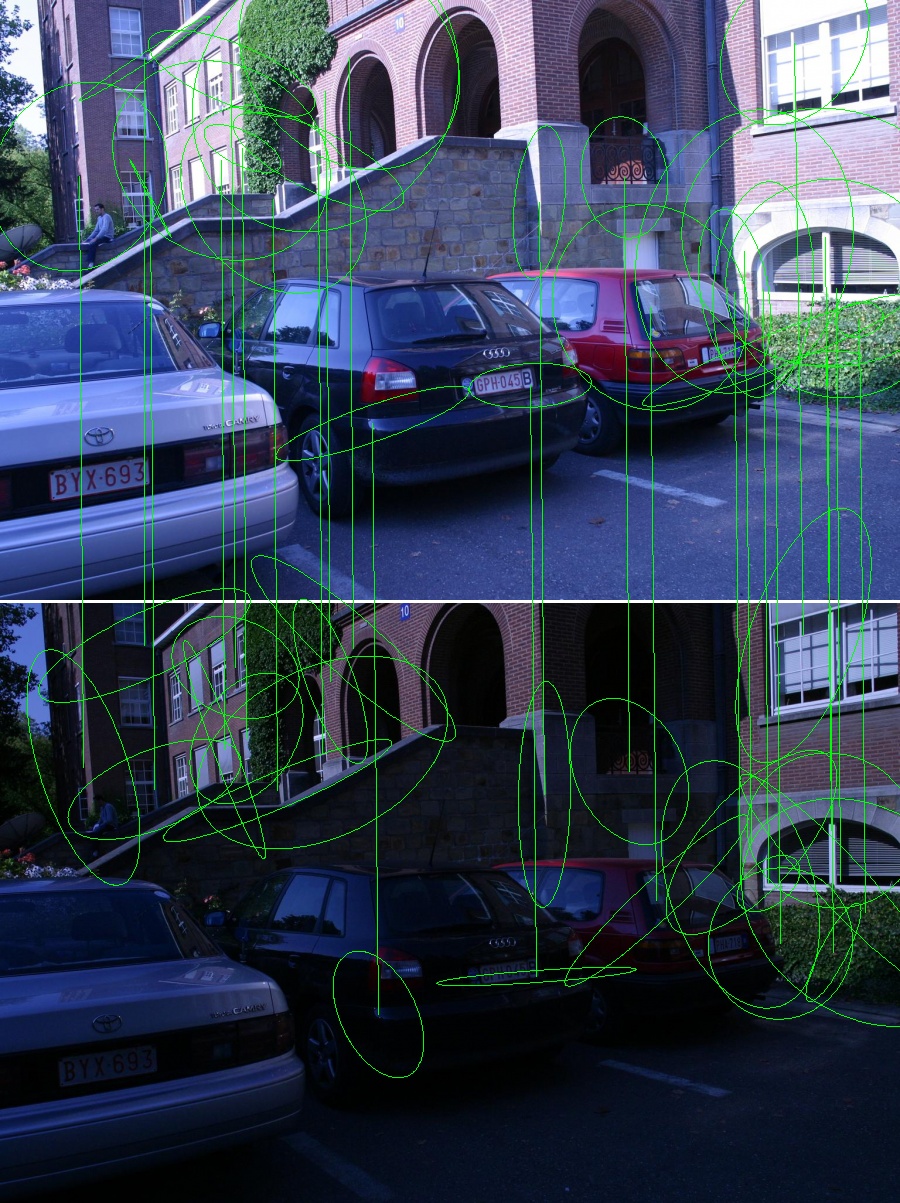}\\
            \vspace{0.2cm}
            \includegraphics[width=\linewidth]{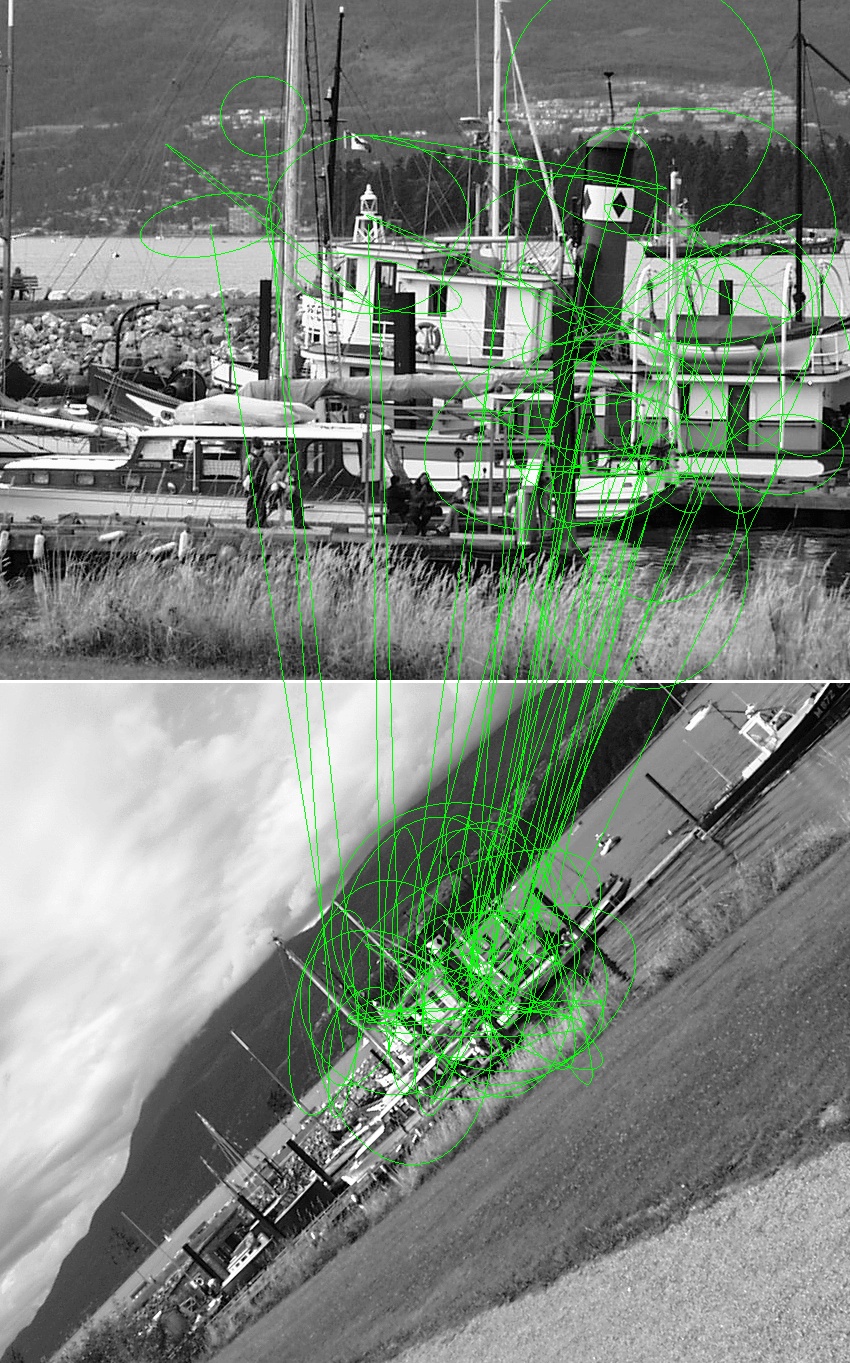}\\
        \end{minipage}
    }
    \vspace{0.02cm}
    \subfloat[Ours]{
        \begin{minipage}[b]{0.2\linewidth}
            \centering
            \includegraphics[width=\linewidth]{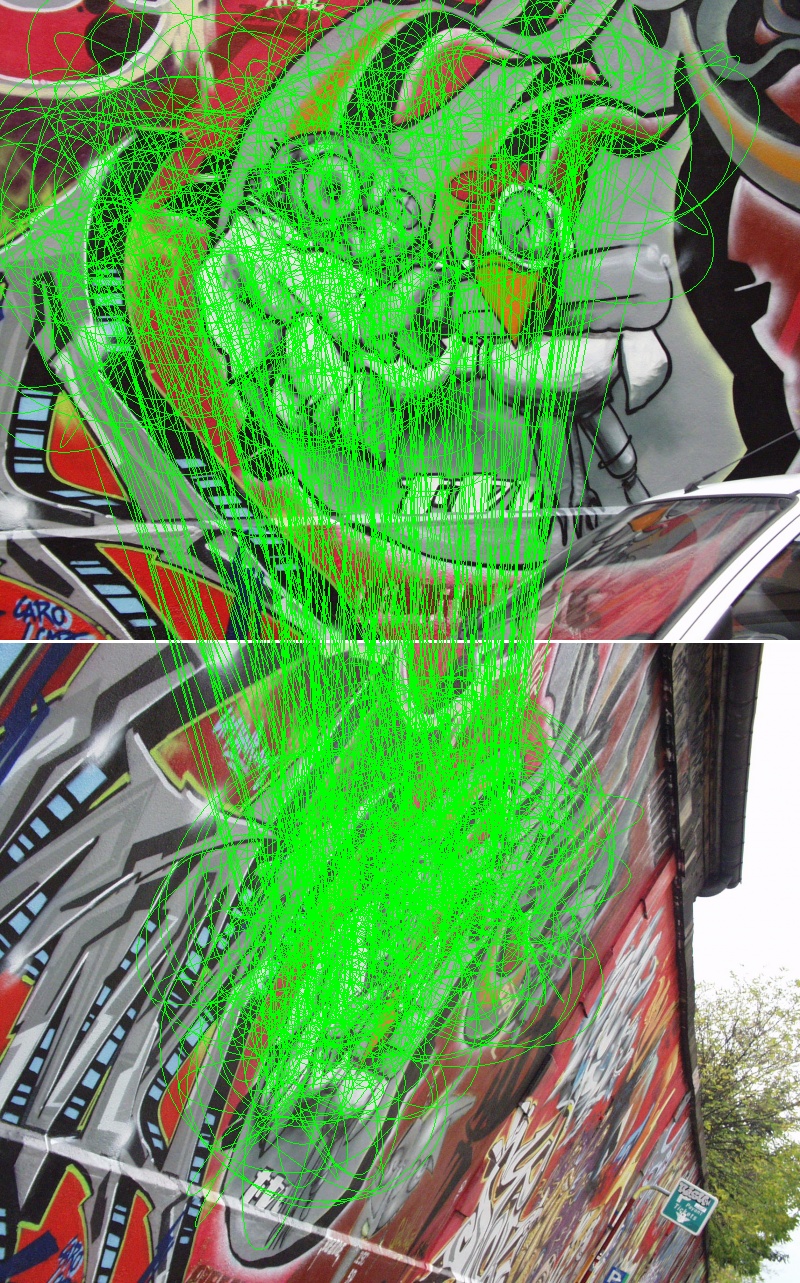}\\
            \vspace{0.2cm}
            \includegraphics[width=\linewidth]{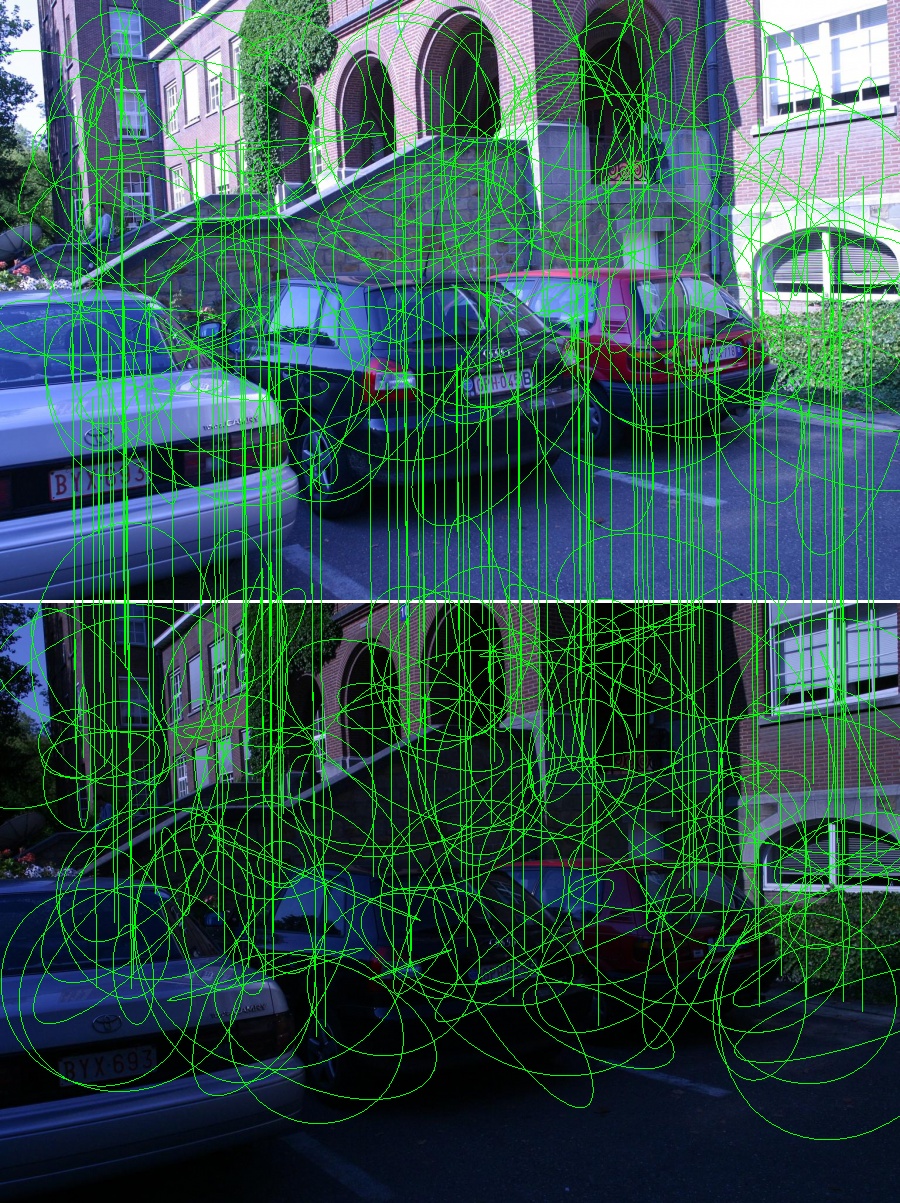}\\
            \vspace{0.2cm}
            \includegraphics[width=\linewidth]{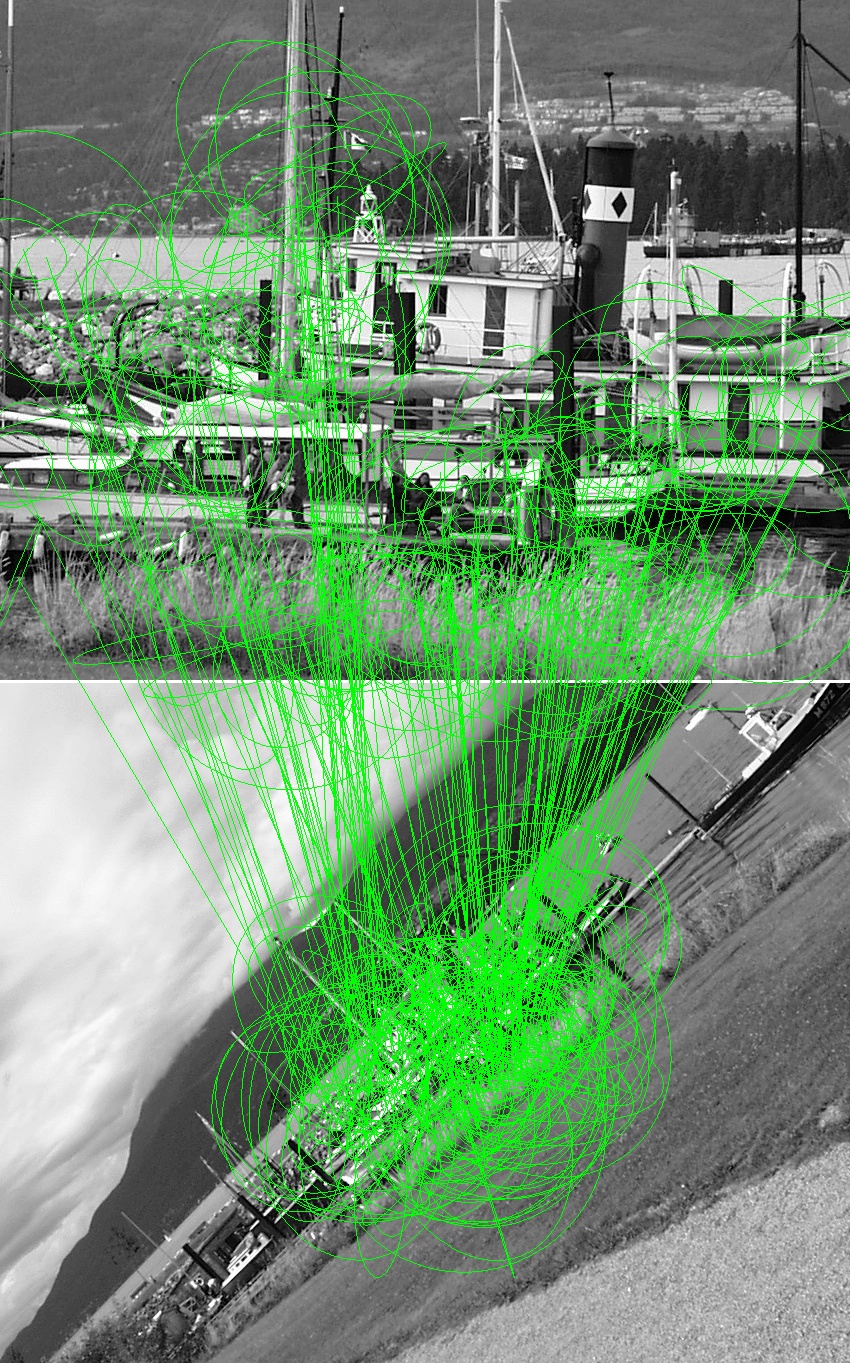}\\
        \end{minipage}
    }
    \centering
    \caption{The image matching results in the Extreme View Dataset \cite{MISHKIN201581}. Our method could finds the highest number of correct matches.}
\label{evd}
\end{figure*}

\begin{equation}
\label{sdF_pc2}
    SD_P({E_{PC}} )=\frac{Z_0^2}{Z_1^2+Z_2^2+Z_3^2+Z_4^2},
\end{equation}
\noindent where
\begin{equation}
\label{sdF_pc1}
    \left\{
    \begin{array}{ll}
        Z_0 = x_1 (f_{31}+f_{11} x_2+f_{21}  y_2)+ y_1 (f_{32}+\\f_{12}x_2+f_{22}y_2) +f_{13}x_2+f_{23} y_2+ f_{33},\\
        Z_1  = f_{31}+ f_{11}  x_2+ f_{21} y_2, \\Z_2  =f_{32}+ f_{12}  x_2+ f_{22}  y_2,\\
        Z_3  =f_{13}+ f_{11} { x_1}+ f_{12}  y_1,\\ Z_4 =f_{23}+f_{22}y_1+f_{21} x_1,          
    \end{array}
    \right.
\end{equation}
\noindent the $f_{i j}$, ($ i, j \in{\{1, 2, 3\}}$) is an element in the fundamental matrix.
The affine transformation constraints is as follows:
\begin{equation}
S D_A\left(E_{A C}\right)_{(1: 2)}=S D_A\left( A ^{-T}\left( F ^T p _2\right)_{(1: 2)}+\left( F p _1\right)_{(1: 2)}\right) .
\label{ac_con}
\end{equation}
When $G(X)$ is the constraint on the first row in Eq.~\ref{ac_con} in the paper. 
We take partial derivatives of $x_1$\dots $a_{22}$. 
Let $M_0 = G(X)$. The remaining terms are $M_1=\frac{\partial M_0}{\partial a_{11}} $, $ M_2=\frac{\partial M_0}{\partial y_1 } $, $ M_3=\frac{\partial M_0}{\partial x_2 }$, $M_4=\frac{\partial M_0}{\partial a_{21}} $, $M_5=\frac{\partial M_0}{\partial x_1 }$, $M_6=\frac{\partial M_0}{\partial y_2 }$. 
The first one can be formulated as follows:
\begin{equation}\label{sd_ac1}
    \begin{split}
        SD_A(E_{AC})_{(1)}=\frac{M_0^2}{M_1^2+M_2^2+M_3^2+M_4^2+M_5^2+M_6^2},\\
    \end{split}
\end{equation}
\noindent where
\begin{small}
    \begin{equation}\label{sd_ac1_more}
        \left\{
        \begin{array}{ll}
            M_0 = x_1(a_{11}f_{11} + a_{21}f_{21}) + y_1(a_{11}f_{12} +  a_{21}f_{22})\\  + a_{11}f_{13} + a_{21}f_{23} + f_{11}x_2 + f_{21}y_2 + f_{31},\\
            M_1  = f_{13} + f_{11}x_1 + f_{12}y_1, \\ M_2  =a_{11}f_{12} + a_{21}f_{22}, \qquad
            \\ M_3 = f_{11}, \\ M_4 =f_{23} + f_{21}x_1 + f_{22}y_1,\\ 
            M_5  =a_{11}f_{11} + a_{21}f_{21},\\ M_6 =f_{21},\\         
        \end{array}
        \right.
    \end{equation}
\end{small}
Similarly, the second one is formulated as
\begin{equation}\label{sd_ac2}
    \begin{split}
        SD_A(E_{AC})_{(2)}=\frac{N_0^2}{N_1^2+N_2^2+N_3^2+N_4^2+N_5^2+N_6^2},\\
    \end{split}
\end{equation}
where 
\begin{small}
\begin{equation}\label{sd_ac2_more}
    \left\{
    \begin{array}{ll}
        N_0 =x_1(a _{12}f_{11} + a_{22}f_{21}) + y_1(a _{12}f_{12} + a_{22}f_{22}) \\+ a _{12}f_{13} + a_{22}f_{23} + f_{12x_2} + f_{22}y_2 + f_{32} ,\\
        N_1  =f_{13} + f_{11}x_1 + f_{12}y_1 ,\\ N_2  = a _{12}f_{11} + a_{22}f_{21} ,\\  N_3  = f_{12},\\  
        N_4  =f_{23} + f_{21}x_1 + f_{22}y_1 ,\\ N_5 = a_{12}f_{12} + a_{22}f_{22}, \\  N_6 =f_{22}, 
    \end{array}
    \right.
\end{equation}
\end{small}

\begin{figure*}[t]
\vspace{-0.5cm}
    \centering
    \subfloat[VLFeat]{
        \begin{minipage}[b]{0.23\linewidth}
            \centering
            \includegraphics[width=\linewidth]{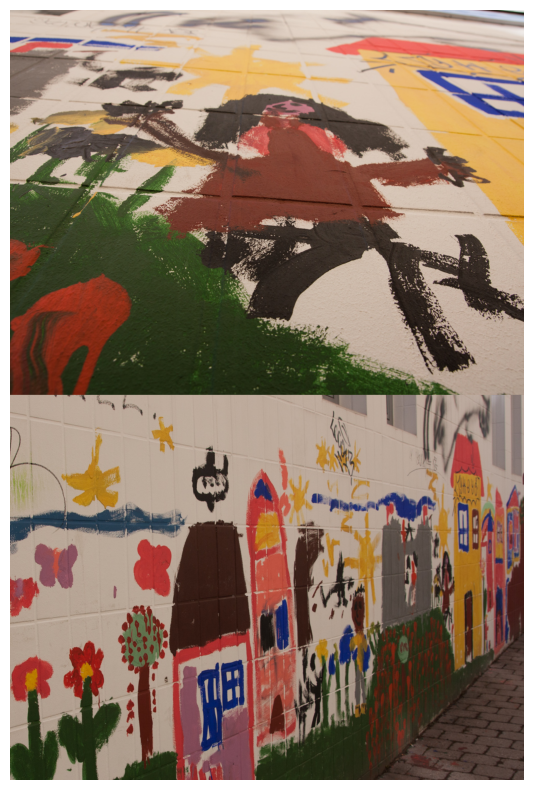}\\
        \end{minipage}
    }
    \subfloat[AffNet]{
        \begin{minipage}[b]{0.23\linewidth}
            \centering
            \includegraphics[width=\linewidth]{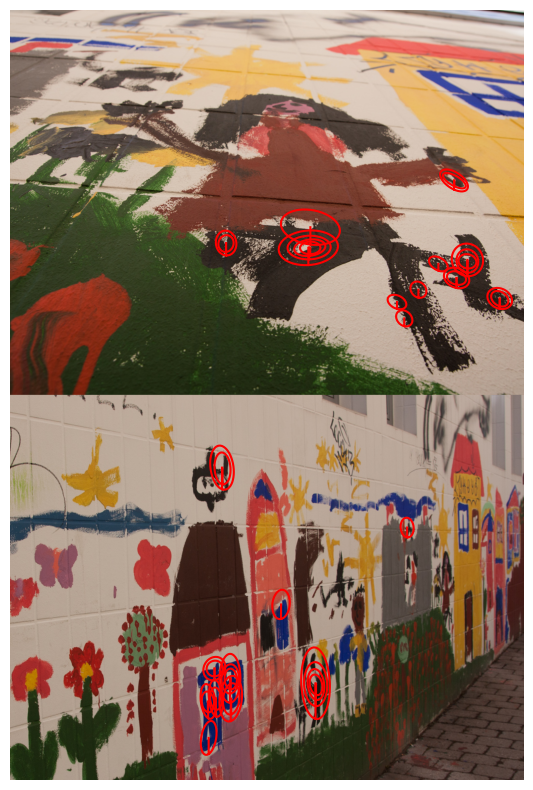}\\
        \end{minipage}
    }
    \subfloat[ASIFT]{
        \begin{minipage}[b]{0.23\linewidth}
            \centering
            \includegraphics[width=\linewidth]{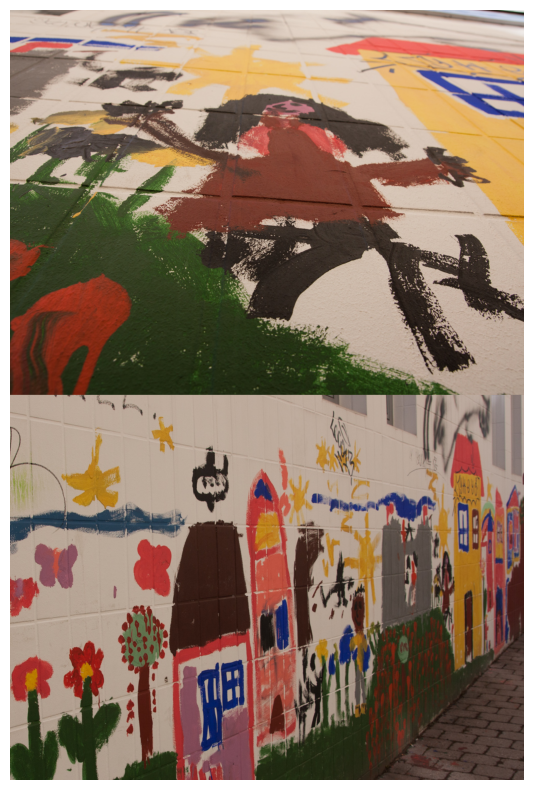}\\
        \end{minipage}
    }
    \subfloat[Ours]{
        \begin{minipage}[b]{0.23\linewidth}
            \centering
            \includegraphics[width=\linewidth]{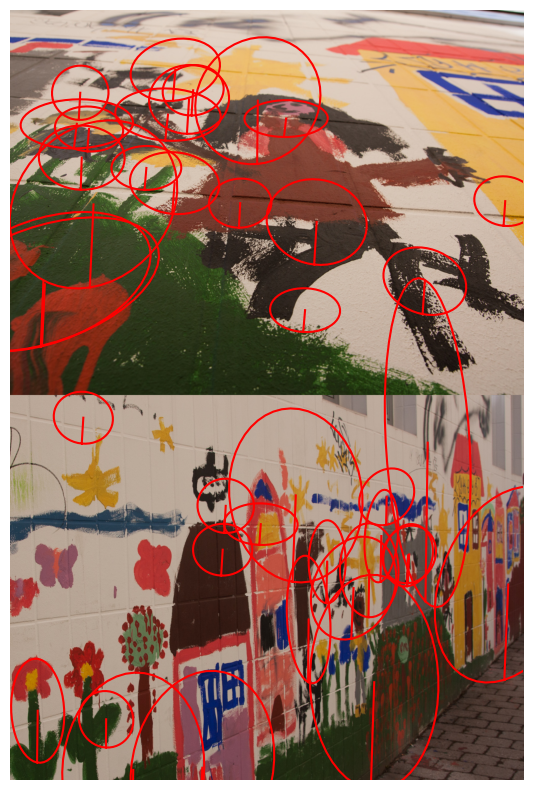}\\
        \end{minipage}
    }
    \centering
    \vspace{-0.3cm}
    \caption{Failure modes. Other methods also fail.}
\label{reb_fg1}
\end{figure*}

\section{Image Matching on EVD}
\label{Image-Matching on EVD}
\noindent\hspace{1em} Affine features are beneficial for matching images with large viewpoint changes because they utilize further 
geometric information compared to their point-based counterparts. 
We now show additional results on the Extreme View dataset~\cite{mishkin2015mods}, whose average viewpoint change is substantially larger than that of the HPatches dataset~\cite{Hptches}. 
The dataset with the ground truth is available on the web-page\footnote{http://cmp.felk.cvut.cz/wbs/index.html}.

\noindent\textbf{Evaluation protocol}.  We compare the proposed method with the view-synthesis-based Affine-SIFT  (ASIFT)~\cite{ASIFT}, the VLFeat library~\cite{Vedaldi2010VlfeatAO}, and the learning-based AffNet~\cite{affnet}. Following the protocol~\cite{affnet}, we report the number of successfully matched image pairs and the average number of correct inliers per matched pair.

\noindent \textbf{Results.} The average inlier numbers and the number of successfully matched image pairs are shown in Table~\ref{tab:5}. Example results are shown in the Fig.~\ref{evd}. Only the correct matches are displayed. Our method has a significant advantage in terms of matching quantity at the same pixel error threshold. Benefiting from the use of dense matching, and through the estimation of affine features, our method obtains more accurate matches in the case of large viewpoint change. This experiment demonstrates that our method is more robust than other affine-based ones to large viewpoint changes. This signifies that the affine correspondences we extract are of better quality. 
This can be attributed to our pipeline design for affine correspondence extraction, leveraging a combination of geometric constraints.
\begin{table}[ht]
\centering
\caption{The comparison of affine extractors on a wide baseline stereo dataset EVD~\cite{MISHKIN201581} following the protocol in~\cite{Mishkin2015WxBSWB}. The number of successfully matched image pairs (N) and the average number of correct inliers (inl.) are presented. The best result is in bold.}
\resizebox{0.8\linewidth}{!}
    {
    \setlength{\tabcolsep}{2mm}{
        \begin{tabular}{l c c c c c}
            \hline
            \multirow[b]{2}{*} &VLFeat  \cite{Vedaldi2010VlfeatAO} & AffNet \cite{affnet} &  ASIFT  \cite{ASIFT} & Ours \\
            \hline
            N. & 2 & 4 & 2 & \pmb{11} \\
            inl. & 56 & 34 & 64 & \pmb{137} \\
            \hline
        \end{tabular}
        }
    }
    \label{tab:5}
\end{table}
\section{Relative Pose Estimation on ScanNet-1500}
\label{ScanNet-1500}
\noindent\hspace{1em} The ScanNet~\cite{ScanNet} is a large-scale indoor dataset that is used to target the task of indoor pose estimation. This dataset is challenging since it contains image pairs with wide baselines and extensive texture-less regions. We follow the evaluation in SuperGlue~\cite{superglue}.
\begin{table}[t]
    \centering
    \caption{Relative pose Accuracy Under the recall Curve (AUC; higher is better) thresholded at $5^\circ$, $10^\circ$, and $20^\circ$ on the ScanNet-1500~\cite{ScanNet}. All methods run RANSAC-based essential matrix estimation, except for the last row, where we run the affine-based GC-RANSAC~\cite{gc-ransac2022,barath_polic_förstner_sattler_pajdla_kukelova_2020}, benefiting from the affine correspondences that we obtain.The best results are in bold.}
    \begin{tabular}{l c c c }
    \hline AUC@ $\rightarrow$ & $5^{\circ} \uparrow$ & $10^{\circ} \uparrow$ & $20^{\circ} \uparrow$ \\
    \hline LoFTR \cite{sun2021loftr} {\tiny$CVPR^{21}$ } & 22.1 & 40.8 & 57.6 \\
    \hline ASpanFormer \cite{Chen2022ASpanFormerDI}  {\tiny$ECCV^{22}$} & 25.6 & 46.0 & 63.3 \\
    \hline PDC-Net+ \cite{PDC-Net+} {\tiny$TPAMI^{23}$} & 20.3 & 39.4 & 57.1 \\
    \hline DKM \cite{DKM}  {\tiny$CVPR^{23}$ } & 29.4 & 50.7 & 68.3 \\
    \hline RoMA\cite{roma}  {\tiny$CVPR^{24}$ } & 31.8 & 53.4 & 70.9 \\
    \hline Ours(RANSAC) & 30.7 & 51.7 & 69.0 \\
    \hline Ours(aff.\ GC-RANSAC) &\pmb{33.1}  & \pmb{55.9} & \pmb{73.4} \\
    \hline
    \end{tabular}
    \label{tab:6}
\end{table}
 	
\noindent\textbf{Evaluation protocol}. 
\noindent Following~\cite{superglue} and~\cite{sun2021loftr}, we report the AUC of the pose error at thresholds $\left(5^{\circ}, 10^{\circ}, 20^{\circ}\right.$). 
To compare with existing methods on the same baseline, we utilize RANSAC as implemented in the OpenCV library to solve for the essential matrix from predicted matches as previous methods do~\cite{sun2021loftr}. 
To demonstrate that the estimated affine frames are beneficial for pose estimation, we also run the affine correspondence-based Graph-Cut RANSAC~\cite{gc-ransac2022,barath_polic_förstner_sattler_pajdla_kukelova_2020}, designed specifically to leverage affine shapes together with the point locations.

\noindent \textbf{Result}.
As shown in Table~\ref{tab:6}, the proposed method with RANSAC achieves good results.
When leveraging the estimated affine shapes with GC-RANSAC, the proposed method achieves the best performance.

\section{Failed Cases}
\label{sec:Failed Cases}
\noindent\hspace{1em} Fig.~\ref{reb_fg1} shows the failure cases caused by significant viewpoint changes and large scale variations. 
However, all other tested baselines fail in these cases.

\appendix

{\small
\bibliographystyle{ieeenat_fullname}
\bibliography{sec/11_references}
}